\documentclass[11pt]{article}
\usepackage[T1]{fontenc}
\usepackage[utf8]{inputenc}
\usepackage{textcomp}
\usepackage{lmodern}
\usepackage[a4paper,margin=1in]{geometry}
\usepackage{amsmath,amssymb}
\usepackage{graphicx}
\usepackage{float}
\usepackage{xcolor}
\usepackage{booktabs,longtable,array,calc,etoolbox}
\usepackage{microtype}
\usepackage{url}
\usepackage[hypertexnames=false,hidelinks]{hyperref}
\usepackage{bookmark}
\providecommand{\real}[1]{#1}

\title{HydroJEV: A one-second, training-free screen for cyber-attack and fault attribution in water distribution networks}
\author{Tianwei Mu$^{a,b,c}$ \and Shengyan Jiang$^{a}$ \and Mingzhe Yuan$^{b,d,*}$ \\ Qing Luo$^{d}$ \and Min Xiao$^{d}$ \and Wenhong Wang$^{b}$ \\ Jun Li$^{a}$ \and Manhong Huang$^{e,*}$}
\date{}
\begin{document}
\maketitle
\begin{center}\small
$^{a}$School of Municipal Engineering and Environment, Shenyang Jianzhu University, Shenyang 110168, China\\
$^{b}$Guangzhou Institute of Industrial Intelligence, Guangzhou 510000, China\\
$^{c}$Shenyang Institute of Automation, Chinese Academy of Sciences, Shenyang 110169, China\\
$^{d}$Key Laboratory of Ecological Restoration of Regional Contaminated Environment, Ministry of Education, College of Environment, Shenyang University, Shenyang 110044, China\\
$^{e}$College of Environmental Science and Engineering, State Environmental Protection Engineering Center for Pollution Treatment and Control in Textile Industry, Donghua University, Shanghai 201620, China\\
Corresponding authors: mzyuan@sia.cn (M. Yuan); huangmanhong@dhu.edu.cn (M. Huang).
\end{center}
\begin{abstract}
When a SCADA alarm is raised in a water distribution network, operators
must decide quickly whether it reflects a cyberattack, a physical fault,
a normal transient or a faulty sensor. Supervised classifiers need
labelled incidents that utilities rarely have, and frontier large
language models (LLMs) take tens of seconds per decision. We tested
whether Jev, a training-free model that returns class probabilities in
about one second, can serve as the first tier of this triage. On a
four-class cause-attribution benchmark built on the C-Town network in
EPANET, Jev was compared with a hand-written rule tree, a supervised
classifier and seven cloud LLMs on identical evidence in four sealed,
pre-registered rounds. With only a label-free prior correction, Jev
matched the rule tree (macro-F1 0.62-0.64 against 0.56-0.61 in
distribution) and exceeded the supervised classifier by 0.36-0.42 on
event subtypes absent from its labels, in all four rounds, and it
outperformed the classifier whenever fewer than about four labelled
events per class were available. Jev also decided 20-40 times faster
than frontier LLMs. Accepting only benign Jev verdicts confirmed by the
rule tree spared an LLM reviewer 35-38\% of windows on fresh sealed sets
without loss of macro-F1. Transferred unchanged to two further networks,
this gated cascade stayed within the non-inferiority margin of its
reviewer on all four sets. A fast, training-free screen can therefore
take over about a third of the review load in SCADA anomaly triage while
preserving the accuracy of deliberate review.
\end{abstract}

\par

\section*{Graphical abstract (TOC graphic)}
\begin{center}
\includegraphics[width=\linewidth,height=0.86\textheight,keepaspectratio]{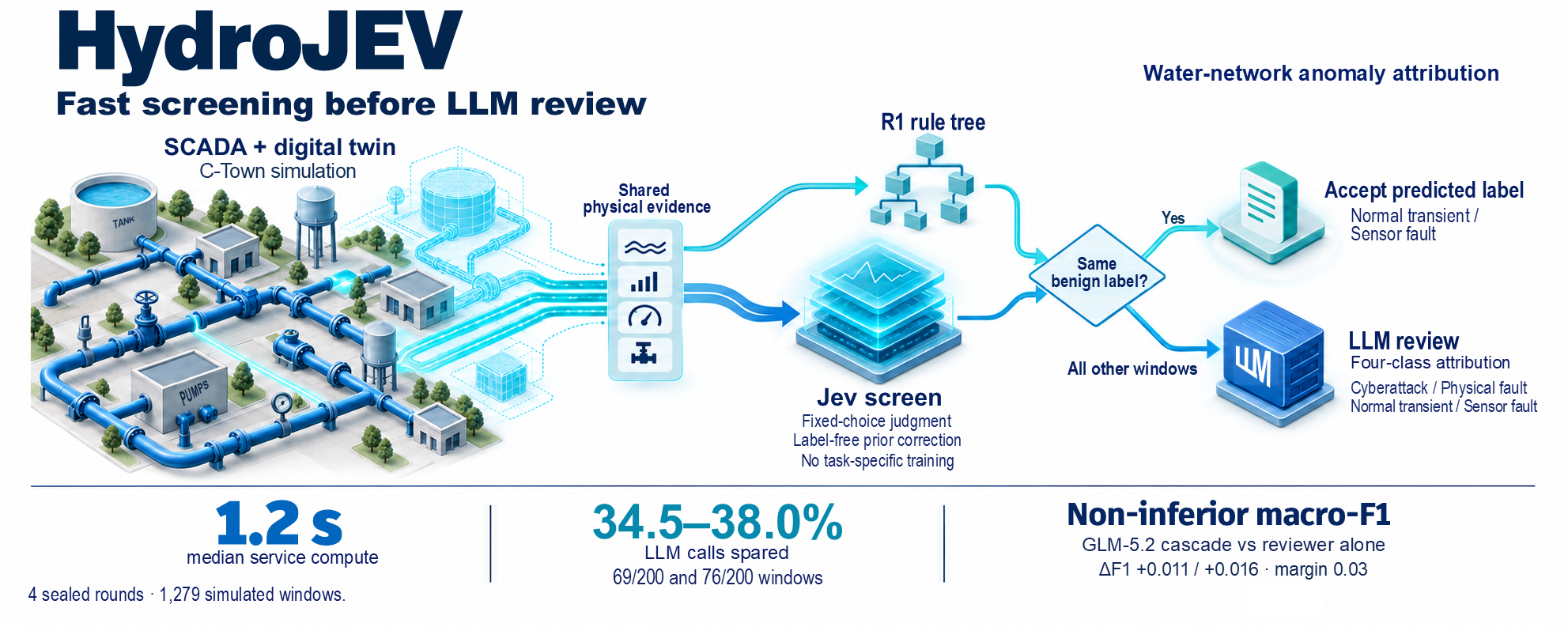}
\end{center}
\noindent\textbf{TOC graphic.} A one-second, training-free typed-judgment screen for anomaly cause attribution: it is fast, robust to event subtypes absent from supervised labels, and spares an LLM reviewer about a third of its calls without loss of accuracy; LLM review remains necessary for concealed attacks.

\section{Introduction}\label{introduction-1}

Water distribution networks (WDNs) are among the most consequential
infrastructures of an urbanizing world (Larsen et al., 2016), and their
operation is being digitized rapidly: supervisory control and data
acquisition (SCADA) systems, dense sensing, hydraulic digital twins and
machine learning now sit between the treatment plant and the tap
(Eggimann et al., 2017; Fu et al., 2022; Richards et al., 2023). The
same connectivity creates a cyber-physical attack surface in which an
adversary can manipulate control logic or falsify the telemetry on which
operators rely (Hassanzadeh et al., 2020; Rasekh et al., 2016).
Detecting that something is abnormal has become routine. Model-based
monitors compare live telemetry with a calibrated hydraulic model(Housh
and Ohar, 2018), and data-driven monitors learn the manifold of normal
operation and score its violation (Abokifa et al., 2019; Chandy et al.,
2019; Taormina and Galelli, 2018); both lineages were developed and
compared on the Battle of the Attack Detection Algorithms (BATADAL) and
its C-Town network (Taormina et al., 2017; Taormina et al., 2018).

An alarm, however, is only the start of the operator's work. The same
residual pattern can arise from a cyberattack on control or telemetry,
from a real hydraulic fault such as a burst, from an unusual but
legitimate operating condition such as a demand surge or a manual pump
switch, or from a single faulty instrument. Each cause calls for a
different response (incident response, a repair crew, no action or a
maintenance ticket), so attributing the cause is a triage decision that
must be made quickly and often, on evidence that is hourly and partial.
Detection benchmarks score whether an attack is present, not which cause
produced the anomaly, and this attribution step is left to the operator.

Three families of tools can support the triage, and each carries a
characteristic cost. Supervised classifiers are accurate when labelled
examples of every event type exist, but labelled cyber incidents are
rare in a utility by construction, and a new attack or fault mechanism
is by definition absent from the training labels. Hand-written rule
trees need no labels and encode the engineer's reasoning transparently,
but they must be engineered by experts and revised for each new
mechanism. General-purpose large language models (LLMs) can reason over
a textual rendering of the evidence without training and are being
explored for decision support in WDNs (Goldstein et al., 2026; Wang et
al., 2026) and as zero-shot anomaly detectors (Alnegheimish et al.,
2024). They generate free text that must be parsed into a decision, a
step exposed to hallucination (Farquhar et al., 2024; Ji et al., 2023)
and frontier reasoning models take tens of seconds to minutes per
decision, which is slow for a control room that must clear many alarms.

Jev, the first of a class of ``System One'' models (TypeSafe, 2026),
instead returns a typed judgment, a probability distribution over a
fixed set of options, in about one second. By analogy with fast,
intuitive human judgment (Kahneman, 2011), such a model is a natural
candidate for the first tier of a pipeline in which fast verdicts settle
the easy cases and a slower, deliberate reviewer handles the rest, the
logic of classifier cascades (Viola and Jones, 2001), selective
classification (Geifman and El-Yaniv, 2017) and LLM cascades (Chen et
al., 2023). Whether a training-free typed-judgment model is accurate
enough for this role in cyberattack and fault attribution, where it
fails, and whether a cascade built on it preserves the accuracy of the
slow reviewer have not been measured. Such a measurement must also guard
against the optimism of methods tuned on the data on which they are
reported, which calls for sealed, pre-registered test sets.

Here we make this measurement on public assets and call the resulting
screen-and-review pipeline HydroJEV. We built a four-class
cause-attribution benchmark on C-Town in EPANET (Klise et al., 2017;
Rossman, 2000) with disjoint families of cyberattack, fault and benign
event subtypes, and evaluated Jev, a hand-written rule tree, a
supervised classifier and seven cloud LLMs on identical evidence in four
sealed, pre-registered rounds.

We asked three questions: (1) whether training-free typed judgment
generalizes where supervision cannot, (2) what the extra accuracy of
frontier LLMs costs in time, and (3) whether an agreement-gated cascade
turns the screen's speed into reviewer savings without loss of accuracy
on fresh sealed sets. We also characterise the conditions of use of the
screen: the role of the agreement gate, the event classes that are
hardest to attribute and the transfer to two further networks.

\section{Data and methods}\label{data-and-methods}

\subsection{Benchmark}\label{benchmark}

Each analysed window is a 6 h SCADA record from a stochastic EPANET
simulation of C-Town (\textbf{Fig. 1a}; \textbf{Fig. S1}) and belongs to
one of four classes defined by mechanism. A cyberattack is concealed
manipulation, in which a reported reading or actuator status is
contradicted by independent, physically related measurements or an
actuator runs contrary to its control logic. A physical fault is a real
hydraulic change whose telemetry is mutually consistent but which the
digital twin cannot explain. A normal transient is honest operation that
departs from nominal, and a sensor fault is a single instrument that
misreports. The subtypes are grouped into four families (\textbf{Table
1}). The reference family supplies the training and development data and
the in-distribution sealed sets. Three unseen families (A, B and C) were
each specified before any of their windows was generated and were used
only for sealed evaluation. Family C is a look-alike family: each class
is paired with a subtype that mimics another class, such as a pump trip,
which is physical but resembles a spoofed stop, or a masked duty swap,
which is an attack but resembles an operator rotation.

\textbf{Table 1.} Event subtypes by class and family.

\begin{longtable}[]{@{}
  >{\raggedright\arraybackslash}p{(\linewidth - 8\tabcolsep) * \real{0.1448}}
  >{\raggedright\arraybackslash}p{(\linewidth - 8\tabcolsep) * \real{0.2501}}
  >{\raggedright\arraybackslash}p{(\linewidth - 8\tabcolsep) * \real{0.1981}}
  >{\raggedright\arraybackslash}p{(\linewidth - 8\tabcolsep) * \real{0.2291}}
  >{\raggedright\arraybackslash}p{(\linewidth - 8\tabcolsep) * \real{0.1780}}@{}}
\toprule

\begin{minipage}[b]{\linewidth}\raggedright
Class
\end{minipage} & \begin{minipage}[b]{\linewidth}\raggedright
Reference

(training, in distribution)
\end{minipage} & \begin{minipage}[b]{\linewidth}\raggedright
Unseen A

(round 1-2)
\end{minipage} & \begin{minipage}[b]{\linewidth}\raggedright
Unseen B

(round 3)
\end{minipage} & \begin{minipage}[b]{\linewidth}\raggedright
Unseen C (rounds 4-6)
\end{minipage} \\
\midrule

\endhead
\bottomrule

\endlastfoot
Cyberattack & Overpump, masked level; pump off, spoofed running & Pump
on, spoofed off; overpump, level replay & Set-point tamper, unmasked;
pump stop, 24 h replay & Level spoof, starve; masked duty swap \\
Physical fault & Burst & Pipe closure & Pump wear; dual leak & Growing
leak; pump trip \\
Normal transient & Demand rise; demand fall; operator switch & Demand
pattern shift & Duty pump rotation; temporary demand peak & Gradual
demand ramp; erratic demand \\
Sensor fault & Stuck; drift; offset & Gain error & Dropout to zero;
intermittent spikes & Noise burst; lagged channel \\
\end{longtable}

The reference family provides the 200 training windows (50 per class),
the 40 development windows and the in-distribution sealed sets; each
unseen family is used only for its own sealed set.

C-Town (388 junctions, 429 pipes, 7 tanks, 11 pumps in five stations, 4
valves and 1 reservoir, with level-based and timed controls) was
simulated unchanged in EPANET 2 (Rossman, 2000) through WNTR (Klise et
al., 2017). Each scenario is a stochastic 48-54 h hourly simulation with
4\% multiplicative demand noise and produces two records: the true
hydraulic state and a reported SCADA historian, which is the true state
plus Gaussian measurement noise with any manipulation or instrument
fault applied on top. The historian carries tank levels, pump and valve
statuses and flows, junction pressures, tank inflow meters and source
outflow meters. An event starts at hour 24-48, and the analysed window
is the 6 h ending 2-6 h after onset. A digital twin, the same EPANET
model with nominal demand, is re-synchronized to the reported tank
levels 11 h before the window end, always before onset, and replays the
reported actuator states. A window is regenerated from a deterministic
seed ladder when its event is not yet physically visible inside it, for
instance when an interlock has not yet been defeated or a faulty sensor
reads within noise of the true value. The seed ladders of different
splits are disjoint (\textbf{Note S1}; \textbf{Table S1}).

\begin{figure}[tbp]
\centering
\includegraphics[width=\linewidth,height=0.78\textheight,keepaspectratio]{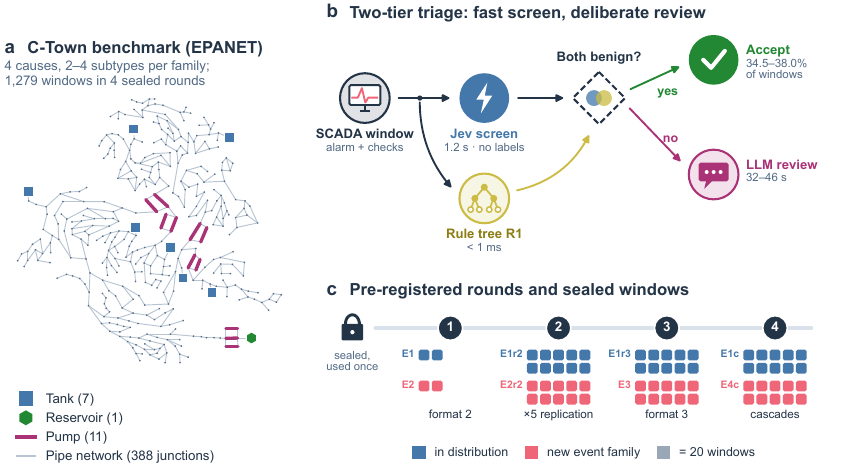}
\caption{Study design. (a) The C-Town benchmark network, with its tanks, reservoir and pumps. Four event classes are instantiated by four disjoint subtype families, and 1,279 windows were evaluated in four sealed rounds. (b) The two-tier triage. Every window is first judged by the training-free Jev screen (median 1.2 s per decision, no labels). A benign Jev verdict that the rule tree R1 confirms is accepted; every other window goes to an LLM reviewer (median time per decision, one request at a time). The ungated rule cascade, which accepts every benign Jev verdict and otherwise uses R1, was also tested. (c) The pre-registered rounds and their sealed sets. Blue, in-distribution (reference-family) sets; red, sets from a new event family. Rounds 5 and 6, which transferred the round-4 screen to two further networks, are described in \textbf{Note S11}.}
\label{fig:1}
\end{figure}

\subsection{Evidence state}\label{evidence-state}

Every method receives the same machine-readable evidence state for a
window. It has six parts: a neutral system description, the control
configuration, the consistency checks, the ten largest twin residuals as
z-scores, a signed summary and a source-meter balance (\textbf{Note
S2}). The ten checks of evidence format 2 compare pump status with flow,
pump head gain and flow with the twin, tank level change with the inflow
meter, tank level, pressure and system demand with the twin, actuator
states with the control logic applied to both reported and twin levels,
and source meters with station pump flows. Each check reports its
status, statistic, threshold and event-free base rate; thresholds are
the 97.5\% quantiles of 80 event-free runs (30 for the source-meter
balance). The signed summary gives the direction and z-score of the
system-demand residual, the spatial pattern and sign of the pressure
deviations, the tank residual signs and the number of corroborating
channels. Evidence format 3, tested in round 3, adds three checks that
use reported values only (\textbf{Note S6}). An automated token scan
verified for every window that no state contained a class name, a
subtype name, a word that names a mechanism or a network name.

\subsection{Jev screen}\label{jev-screen}

Each window was sent once to the Jev service (TypeSafe AI; model
jev-1.13.0 for every call) as a single-choice question over five
options, the four classes and ``insufficient evidence''. The question
gives the class definitions and instructs the model to treat the four
causes as equally likely a priori; it was fixed on the 40 development
windows before any evaluation (\textbf{Note S2}). Jev returns a
probability for each option, to which we apply an inductive, label-free
prior correction in the spirit of batch calibration (Saerens et al.,
2002; Zhou et al., 2024). The prior is Jev's mean answer on the
unlabelled development windows:

\begin{equation}
\pi_{c} = \max\left( 0.01,\ \frac{1}{|\mathcal{D}|}\sum_{w\in\mathcal{D}}^{}p_{c}(w) \right)
\tag{1}
\end{equation}

where \(\pi_{c}\) is the prior of class \(c\), \(\mathcal{D}\) is the
set of 40 unlabelled development windows and \(p_{c}(w)\) is Jev's
probability for class \(c\) on window \(w\). This gives \(\pi\) =
(0.302, 0.429, 0.146, 0.121) for cyberattack, physical fault, normal
transient and sensor fault. The calibrated decision is:

\begin{equation}
{\widehat{y}}_{J}(w) = \arg\max_{c\in\mathcal{C}}\ \frac{\max\left( p_{c}(w),\, 0.005 \right)}{\pi_{c}}
\tag{2}
\end{equation}

where \({\widehat{y}}_{J}(w)\) is Jev's verdict and \(\mathcal{C}\) is
the set of four classes. No label of an evaluated window enters
\textbf{Eqs. 1 and 2}. Raw probabilities (\textbf{Eq. 2} with
\(\pi_{c} = 1\)) and a transductive leave-one-out prior are reported as
ablations (\textbf{Note S4}), and a failed call counts as a wrong
answer.

\subsection{Baselines}\label{baselines}

Jev is compared with a six-branch rule tree (R1), a supervised logistic
regression (R2) and seven general-purpose LLMs (Section 2.5). R1 applies
six ordered rules to the check statuses and the signed summary; it was
fixed before any Jev call and needs no training data (\textbf{Note S2};
\textbf{Table S2}). R1b prepends a physics override to cyberattack based
on the three format-3 checks (\textbf{Note S6}). R2 is a class-balanced
multinomial logistic regression on standardized features from the same
evidence (each check statistic divided by its threshold, the signed
summary quantities and the number of anomalous channels), trained on the
200 labelled reference-family windows. It is evaluated trained on all
labels (R2 full), retrained with each evaluated subtype left out (R2
LOSO), which simulates an event type absent from the labels, and trained
on k labels per class (k = 1, 2, 4, 8, 16, 40; 200 random draws per k),
which gives the label-scarcity curve (\textbf{Note S7}).

\subsection{LLM reviewers}\label{llm-reviewers}

Seven general-purpose cloud LLMs served through Ollama (deepseek-v4-pro,
glm-5.2, gpt-oss:120b, gpt-oss:20b, gemma4, gemma4:31b and minimax-m2.7)
received exactly the same state and question as Jev, at temperature 0
with a fixed seed and a JSON schema returning the five probabilities
(\textbf{Note S3}). Each LLM was calibrated like Jev, with its own mean
answer on the unlabelled development windows in \textbf{Eq. 1}. An
unparseable or zero-mass answer counts as invalid and wrong, and no
request was re-sampled for its content.

\subsection{Screening cascades}\label{screening-cascades}

A screening cascade lets the fast screen decide the windows it can
settle safely and hands the rest to a slower judge (\textbf{Fig. 1b}).
Because the costliest triage error is to dismiss a real attack or fault
as harmless, the screen settles only benign verdicts, \(\mathcal{B}\) =
\{normal transient, sensor fault\}. The rule cascade is:

\begin{equation}
{\widehat{y}}_{rule}(w) = \left\{ \begin{matrix}
{\widehat{y}}_{J}(w), & {\widehat{y}}_{J}(w)\in\mathcal{B} \\
{\widehat{y}}_{R1}(w), & \text{otherwise}
\end{matrix} \right.\
\tag{3}
\end{equation}

where \(\mathcal{B}\) is the set of benign classes and
\({\widehat{y}}_{R1}(w)\) is the verdict of R1. The LLM cascade accepts
a benign Jev verdict only when R1 agrees:

\begin{equation}
{\widehat{y}}_{LLM}(w) = \left\{ \begin{matrix}
{\widehat{y}}_{J}(w), & {\widehat{y}}_{J}(w)\in\mathcal{B}\ \text{and}\ {\widehat{y}}_{J}(w) = {\widehat{y}}_{R1}(w) \\
{\widehat{y}}_{rev}(w), & \text{otherwise}
\end{matrix} \right.\
\tag{4}
\end{equation}

where \({\widehat{y}}_{rev}(w)\) is the calibrated verdict of the
reviewer LLM, glm-5.2 as the primary reviewer and deepseek-v4-pro as the
secondary one. The first case of \textbf{Eq. 3} is the rule gate and the
first case of \textbf{Eq. 4} the agreement gate. A failed Jev call is
never benign and is always reviewed. Both cascades are cost-sensitive
selective classifiers (Geifman and El-Yaniv, 2017) and instances of
model cascading (Chen et al., 2023; Viola and Jones, 2001). Their design
and confirmation criterion are given in Section 2.7 and \textbf{Note
S10}.

\subsection{Sealed evaluation}\label{sealed-evaluation}

The study ran as a sequence of sealed rounds under a written,
pre-registered protocol (\textbf{Table 2}; \textbf{Fig. 1c}). Each
round's design (question, calibration prior, analysis code and
hypotheses) was frozen, with cryptographic hashes of the question, the
source code and the one-shot analysis script, before its sealed windows
were generated. Each sealed set was answered and analysed exactly once
by an analysis script that refuses to overwrite its output, and sets
consumed by one round served only as development data in later rounds.

\textbf{Table 2.} Sealed evaluation rounds.

\begin{longtable}[]{@{}
  >{\raggedright\arraybackslash}p{(\linewidth - 8\tabcolsep) * \real{0.0874}}
  >{\raggedright\arraybackslash}p{(\linewidth - 8\tabcolsep) * \real{0.2732}}
  >{\raggedright\arraybackslash}p{(\linewidth - 8\tabcolsep) * \real{0.2675}}
  >{\raggedright\arraybackslash}p{(\linewidth - 8\tabcolsep) * \real{0.2550}}
  >{\raggedright\arraybackslash}p{(\linewidth - 8\tabcolsep) * \real{0.1169}}@{}}
\toprule

\begin{minipage}[b]{\linewidth}\raggedright
Round
\end{minipage} & \begin{minipage}[b]{\linewidth}\raggedright
Sealed sets (windows)
\end{minipage} & \begin{minipage}[b]{\linewidth}\raggedright
Tested design
\end{minipage} & \begin{minipage}[b]{\linewidth}\raggedright
Design exposure
\end{minipage} & \begin{minipage}[b]{\linewidth}\raggedright
Primary tests
\end{minipage} \\
\midrule

\endhead
\bottomrule

\endlastfoot
1 & E1, in distribution (40); E2, unseen A (40) & Jev, evidence format 2
& training, development & 4 \\
2 & E1r2, in distribution (200); E2r2, unseen A (200) & replication & +
round 1 & 3 \\
2-LLM & E1r2 and E2r2 answered by seven LLMs & same state and question &
+ round-2 Jev results & 14

(two-sided) \\
3 & E1r3, in distribution (199); E3, unseen B (200) & evidence format 3,
physics rules & + rounds 1-2 & 5 \\
4 & E1c, in distribution (200); E4c, unseen C (200) & screening cascades
& cascades designed on round 2, confirmed on round 3 & 7 \\
5 & Net3: N3-V1, N3-N4 (200 each) & round-4 screen, transferred
unchanged & C-Town only & 4 \\
6 & Net1: N1-V1, N1-N4 (200 each) & round-4 screen, transferred
unchanged & + round 5 & 4 \\
\end{longtable}

Sealed sets contain 50 windows per class (10 per class in round 1).
Design exposure lists the data visible when the round was designed. All
primary tests are Holm-corrected within their round; round 5 keeps its
pre-registered family of eight tests, which is conservative
(\textbf{Note S11}).

The cascades were designed on the round-2 sets. Before any confirmation
call, the criterion was fixed that on both round-3 sets the glm-5.2
cascade must lose no more than 0.03 macro-F1 relative to glm-5.2 alone
and accept at least 15\% of windows; only after it was met were the
round-4 sets generated from seed ladders reserved in advance, E4c from
the look-alike family C, which no design step had seen. The primary
metric is four-class macro-F1 (\textbf{Eq. 5}), and every primary test
is compared by a paired bootstrap and Holm-corrected within its round
(\textbf{Eqs. 6 and 7}); the pre-registered primary tests of each round
are listed in \textbf{Table S3}.

\subsection{Statistics}\label{statistics}

The primary metric is four-class macro-F1:

\begin{equation}
F_{1} = \frac{1}{|\mathcal{C}|}\sum_{c\in\mathcal{C}}^{}\frac{2P_{c}R_{c}}{P_{c} + R_{c}}
\tag{5}
\end{equation}

where \(P_{c}\) and \(R_{c}\) are the precision and recall of class
\(c\). Two methods A and B are compared on the same windows with a
class-stratified paired bootstrap of \(B\) = 10,000 resamples, which
gives the replicates \(\Delta_{b} = F_{1,A}^{(b)} - F_{1,B}^{(b)}\) of
the macro-F1 difference. The one-sided superiority and non-inferiority
p-values are:

\begin{equation}
p_{\sup} = \frac{1}{B}\sum_{b = 1}^{B}\mathbf{1}\left[ \Delta_{b} \leq 0 \right],\quad\quad p_{NI} = \frac{1}{B}\sum_{b = 1}^{B}\mathbf{1}\left[ \Delta_{b} \leq - \delta \right]
\tag{6}
\end{equation}

where \(\mathbf{1}\lbrack \cdot \rbrack\) is the indicator function and
\(\delta\) = 0.03 is the non-inferiority margin, used for the LLM
cascade against its reviewer. The two-sided LLM comparisons of round 2
double the smaller tail. Holm's procedure (Holm, 1979) controls the
family-wise error rate at \(\alpha\) = 0.05 within each round's primary
family: with the \(m\) p-values ordered as
\(p_{(1)} \leq \ldots \leq p_{(m)}\), the hypothesis of rank \(i\) is
rejected when:

\begin{equation}
p_{(j)} \leq \frac{\alpha}{m - j + 1}\quad\text{for all}\ j \leq i
\tag{7}
\end{equation}

where \(\alpha\) is the family-wise error rate and \(m\) the number of
primary tests in the round. Proportions carry Wilson 95\% intervals
(Wilson, 1927). Secondary results are uncorrected and are not used for
claims. The label-scarcity crossover is:

\begin{equation}
k^{*} = \min\left\{ k\in\mathcal{K}:\ \frac{1}{D}\sum_{d = 1}^{D}F_{1,R2}^{(k,d)} > F_{1,J} \right\}
\tag{8}
\end{equation}

where \(\mathcal{K}\) = \{1, 2, 4, 8, 16, 40\} is the set of label
budgets per class, \(D\) = 200 is the number of random label draws and
\(F_{1,R2}^{(k,d)}\) is the macro-F1 of R2 trained on draw \(d\) with
\(k\) labels per class.

\subsection{Timing}\label{timing}

Jev time is the service-reported compute time of each call; the client
wall-clock time of the call is also reported (\textbf{Note S9}). In
round 4, a class-stratified subset of 20 windows per sealed set was sent
to each reviewer strictly one request at a time, before any other
round-4 request, so that its time excludes client-side queueing; rounds
5 and 6 repeated this protocol. Round-2 and round-3 LLM times were
recorded under concurrent load and are reported only descriptively. On
the timed subset, the time of a cascade decision is

\begin{equation}
t_{casc}(w) = t_{J}(w) + \mathbf{1}\left[ w\ \text{deferred} \right]\, t_{rev}(w)
\tag{9}
\end{equation}

where \(t_{J}(w)\) and \(t_{rev}(w)\) are the Jev and reviewer times for
window \(w\); R1 runs in under a millisecond and is neglected.

\subsection{Transfer networks}\label{transfer-networks}

Rounds 5 and 6 applied the round-4 screen and cascades unchanged to
EPANET Net3 (92 junctions, 117 pipes, 3 tanks, 2 reservoirs, 2 pumps and
a level-controlled pressure-zone bypass) and EPANET Net1 (9 junctions,
12 pipes, 1 tank, 1 reservoir and 1 level-controlled pump; \textbf{Fig.
S7}). Net1 was used with its demand patterns resampled to a 1 h step,
matching the hourly time base of the event generator; its event-free
hydraulics are unchanged. The question, the evidence format, the Jev and
reviewer priors, R1 and R2 fitted on the C-Town labels (``R2 transfer'')
were frozen from C-Town; only network properties (channel set, check and
residual thresholds from event-free runs, station balance calibration)
were recomputed with the frozen code. The number of monitored junctions
is \(min(8,\ \lfloor n/2\rfloor)\), where n is the number of
non-pump-end junctions with a median pressure above 15 m, which gives 8
on C-Town and Net3 and 4 on Net1. When a network could not host a
subtype, or an event tripped no check inside its window, the seed ladder
drew again, with up to 400 draws per window. Each network received a
reference-family and a family-C sealed set of 200 windows. Per set, two
primary tests were pre-registered, non-inferiority of the glm-5.2
cascade to glm-5.2 alone and superiority of Jev over R2 transfer, and
they were Holm-corrected within their round (\textbf{Note S11}).

\section{Results}\label{results}

\subsection{Accuracy without
training}\label{accuracy-without-training}

Across the four in-distribution sealed sets, calibrated Jev reached a
macro-F1 of 0.620, 0.628, 0.628 and 0.644 (mean 0.63), and the rule tree
R1 reached 0.557, 0.606, 0.602 and 0.590 (mean 0.59; \textbf{Fig. 2a}).
Jev was numerically ahead in every round, but no single difference to R1
was significant, and over all eight sealed sets the paired difference
averaged +0.03 (\textbf{Fig. 2b}). On the four new-family sets the two
were also statistically indistinguishable (mean 0.51 against 0.50). Jev
thus reached the accuracy of an expert-engineered rule tree with neither
rule engineering nor labels. A supervised classifier trained on 200
labelled windows of the same subtypes was clearly better in distribution
(R2 0.887-0.934, mean 0.91), which was pre-registered as expected and is
not a claim of the study. The label-free prior correction was essential
to Jev's accuracy (Section 3.6).

\begin{figure}[tbp]
\centering
\includegraphics[width=\linewidth,height=0.78\textheight,keepaspectratio]{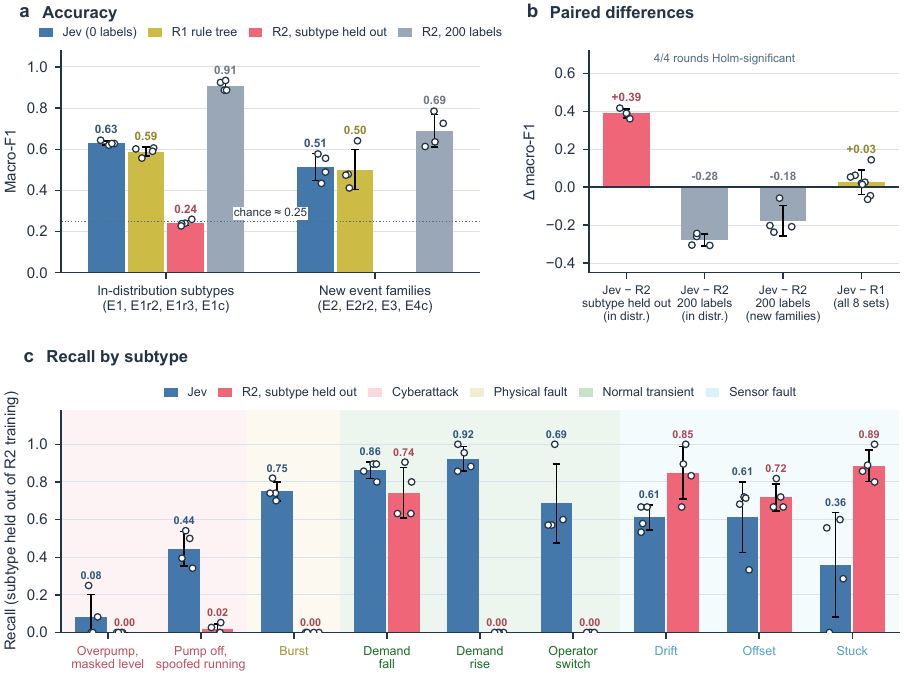}
\caption{Accuracy of training-free and supervised methods on the eight C-Town sealed sets. (a) Macro-F1 of calibrated Jev, the rule tree R1, the supervised classifier R2 retrained with the evaluated subtype held out, and R2 trained on all 200 labels, for the four in-distribution sets (E1, E1r2, E1r3, E1c) and the four new-family sets (E2, E2r2, E3, E4c). Holding out a subtype is defined only for in-distribution sets. The dotted line marks the macro-F1 of a chance-level four-class classifier. (b) Paired differences in macro-F1 per sealed set; the difference to R2 with the subtype held out was a primary test in all four rounds and significant after Holm correction in each. (c) Recall by subtype over the four in-distribution sets, for Jev and for R2 retrained without that subtype; shading gives the true class. In all panels, bars show the mean over sealed sets, error bars \ensuremath{\pm}1 s.d. and white dots the individual sealed sets; the mean is printed above each bar.}
\label{fig:2}
\end{figure}

\subsection{Unseen event subtypes}\label{unseen-event-subtypes}

The strongest and most stable result is the leave-one-subtype-out
comparison, which asks what happens when the event in front of the
operator belongs to a subtype that no training label covers. Jev needs
no retraining and its predictions are unchanged, whereas R2 retrained
without the evaluated subtype collapsed to 0.259, 0.240, 0.238 and
0.227, close to chance. Jev exceeded it by +0.361 {[}95\% CI 0.189,
0.523{]} in round 1, +0.387 {[}0.318, 0.455{]} in round 2, +0.390
{[}0.321, 0.458{]} in round 3 and +0.417 {[}0.344, 0.488{]} in round 4
(all p \textless{} 10\textsuperscript{\ensuremath{-}4}; \textbf{Fig. 2b};
\textbf{Table S3}). All four were primary tests and survived Holm
correction, and the largest margin came from the round-4 set, which was
generated only after the cascade design had been frozen.

The mechanism is visible subtype by subtype (\textbf{Fig. 2c};
\textbf{Note S5}; \textbf{Figs. S2 and S3}). Once a subtype was
withheld, R2 recalled none of the bursts, demand rises, demand falls or
operator switches and almost none of the concealed pump attacks, because
a linear decision surface fitted to the remaining subtypes does not
extend to a new mechanism. Jev, which reads the same evidence against
class definitions rather than against remembered examples, recalled on
average 0.75 of the bursts, 0.92 of the demand rises and 0.86 of the
demand falls. Only on sensor drifts, offsets and stuck sensors, whose
signature resembles the remaining sensor subtypes, did R2 keep its
recall.

Genuinely new event families are harder for every method, and the
advantage does not extend to them. On the new-family sets, fully
supervised R2 remained the most accurate method (mean 0.69 against Jev's
0.51; \textbf{Fig. 2a,b}). On unseen family B, however, R2 failed
completely on one new physical subtype, pump wear (recall 0.00, mistaken
for an attack), of which Jev recalled 0.65 (\textbf{Fig. S2}), so the
two methods fail on different windows.

\subsection{Label requirements}\label{label-requirements}

To ask how many labels a supervised model needs to overtake Jev, we
subsampled R2's training labels to k per class and repeated each k 200
times (\textbf{Fig. 3}; \textbf{Note S7}). On the in-distribution sets,
R2 averaged 0.46 with one label per class and 0.59 with two, against
Jev's 0.63, and overtook Jev only at four (0.72; \textbf{Fig. 3a}). On
the new-family sets it needed four to eight (\textbf{Fig. 3b}). The
crossover k*, the smallest k at which the mean of R2 exceeds Jev
(\textbf{Eq. 8}), was 4 on seven of the eight sealed sets and 40 on E2.
With one label per class, R2 fell below Jev in 96-97\% of draws, and
with two still in 64\% (in distribution) and 82\% (new families) of
draws (\textbf{Fig. 3c}). For a utility, four labelled events of every
cause means four confirmed cyber incidents of each kind, more than most
utilities will ever own and none for a new attack type.

\begin{figure}[tbp]
\centering
\includegraphics[width=\linewidth,height=0.78\textheight,keepaspectratio]{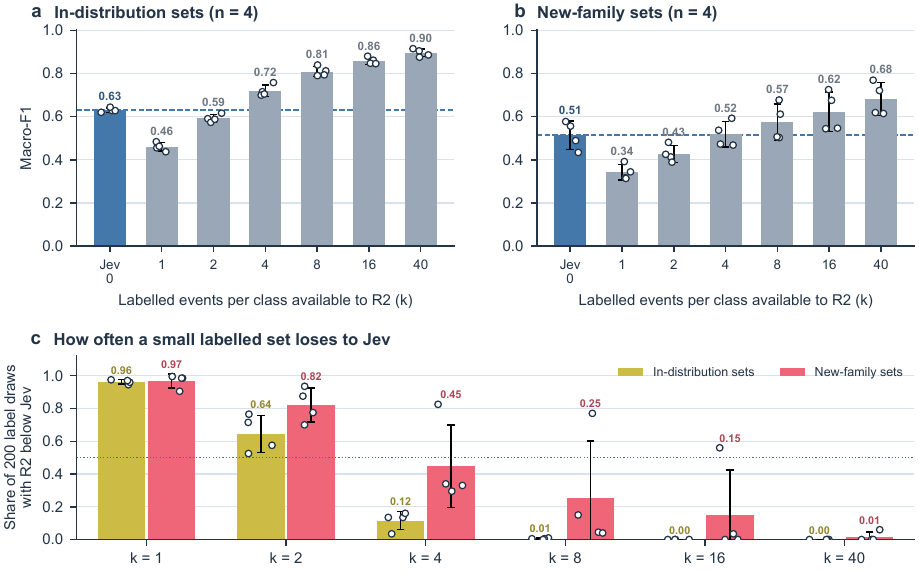}
\caption{Label scarcity. (a, b) Macro-F1 of R2 trained on k labelled events per class, against Jev, which uses no labels, on the four in-distribution sets (a) and the four new-family sets (b). For each set, the R2 value at a given k is the mean over 200 random label draws; the dashed line is Jev's mean. (c) Share of the 200 label draws in which R2 scored below Jev, for each k; the dotted line marks one half. Bars show the mean over the four sealed sets, error bars \ensuremath{\pm}1 s.d. and white dots the individual sets; the mean is printed above each bar. The range over the 200 draws is shown per set in \textbf{Note S7}.}
\label{fig:3}
\end{figure}

\subsection{Comparison with LLMs}\label{comparison-with-llms}

The seven LLMs received the same state and question as Jev, were
calibrated in the same label-free way, and answered all 400 round-2
windows with no invalid answer (\textbf{Fig. 4}; \textbf{Note S3}). Two
frontier reasoning models were significantly more accurate: glm-5.2
scored 0.706 and 0.697 and deepseek-v4-pro 0.722 and 0.637 on E1r2 and
E2r2, against Jev's 0.628 and 0.434 (\textbf{Fig. 4a}). In all, four of
14 Holm-corrected comparisons were significant, all in favour of an LLM
(\textbf{Table S3}); Jev was statistically indistinguishable from
gemma4, gemma4:31b and minimax-m2.7 on both sets and from gpt-oss:120b
in distribution.

The frontier models paid for this accuracy in time. Over 2,722 calls,
Jev's service-reported compute time had a median of 1.22 s (10th-90th
percentile 1.15-1.46 s), and its round-4 calls gave 1.18 s (\textbf{Note
S9}). Sent strictly one request at a time, glm-5.2 took a median of
30.6-31.7 s and deepseek-v4-pro 39.3-46.4 s per decision (means 43.5 and
48.0 s; \textbf{Fig. 4b}), because they generated about 2,300-2,900 and
5,600-8,600 output tokens per decision. A frontier LLM thus needed 20-40
times more time per decision than Jev without queueing, and two to three
orders of magnitude more under the concurrent load of a busy service.

\begin{figure}[tbp]
\centering
\includegraphics[width=\linewidth,height=0.78\textheight,keepaspectratio]{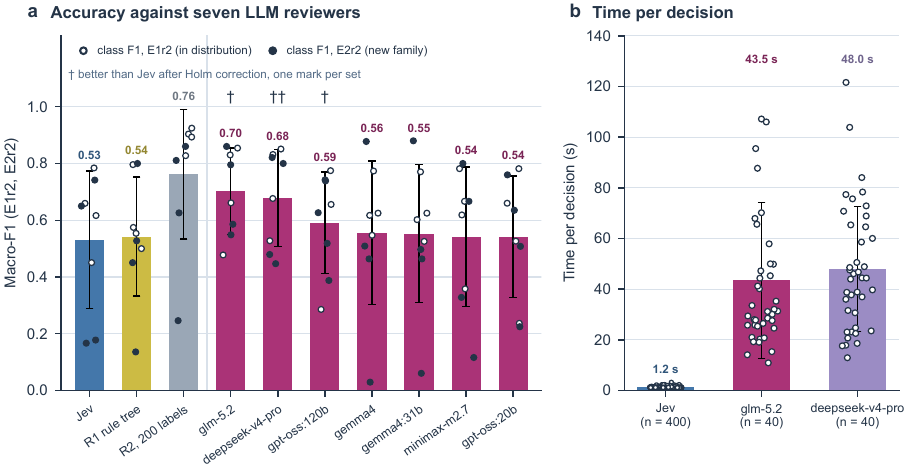}
\caption{Jev against seven general-purpose LLMs, with the same evidence, question and label-free calibration for every model. (a) Macro-F1 on the two round-2 sealed sets (E1r2, E2r2), with the rule tree R1 and R2 trained on 200 labels as references; \textdagger{} marks a set on which the LLM was better than Jev after Holm correction over 14 two-sided tests. With two sets per method, the s.d. only indicates their spread. (b) Time per decision: every round-4 Jev call (n = 400) and the 40 round-4 reviewer decisions per model (20 per set), sent strictly one request at a time. Bars show the mean, error bars \ensuremath{\pm}1 s.d. and white dots the individual sets (a) or decisions (b); the mean is printed above each bar.}
\label{fig:4}
\end{figure}

\subsection{Screening cascades}\label{screening-cascades-1}

Jev's benign verdicts are its most reliable ones: on the four design
sets of rounds 2 and 3, 34-49\% of windows received a benign Jev
verdict, of which 74-86\% were correct. The two cascades of Section 2.6
exploit this asymmetry (\textbf{Eqs. 3 and 4}). Both were designed on
the round-2 sets and checked against their confirmation criterion on the
round-3 sets, and the design sets favoured both (\textbf{Fig. 5a}, light
bars; \textbf{Note S10}).

On the fresh round-4 sets, the LLM cascade was non-inferior to its
reviewer. With glm-5.2, the difference to the reviewer alone was +0.011
{[}\ensuremath{-}0.011, 0.033{]} in distribution and +0.016 {[}0.003, 0.033{]} on the
unseen family, both inside the 0.03 margin after Holm correction (p
\textless{} 10\textsuperscript{\ensuremath{-}4}; \textbf{Fig. 5a}, dark bars); with
deepseek-v4-pro the differences were +0.010 {[}\ensuremath{-}0.016, 0.035{]} and
\ensuremath{-}0.009 {[}\ensuremath{-}0.038, 0.018{]} (\textbf{Note S12}; \textbf{Fig. S13}). The
agreement gate settled 69 and 76 of the 200 windows (34.5\% and 38.0\%),
so the reviewer was needed for 62-66\% of them, and on the timed windows
the mean time per decision (\textbf{Eq. 9}) fell from 43.5 to 30.6 s
with glm-5.2 and from 48.0 to 34.2 s with deepseek-v4-pro (\textbf{Fig.
5d}).

The ungated rule cascade performed on a par with its components on the
fresh sets, so none of its four superiority tests was met. It gained
+0.047 over R1 and +0.049 over Jev on the design sets but \ensuremath{-}0.001 and
\ensuremath{-}0.035 on the fresh ones (\textbf{Fig. 5a}). The difference lies in the
physical class. Jev called 13 of the 50 E1c bursts benign, and because
the rule cascade accepts every benign Jev verdict without a second
opinion, burst recall fell from R1's 0.78 to 0.52 (\textbf{Fig. S5}).
Across the eight C-Town sealed sets, requiring R1's agreement lowered
the accept rate from 0.45 to 0.29, raised the accept precision from 0.75
to 0.83 (\textbf{Fig. 5b}) and cut accepted physical faults from 8.4 to
1.9 per set (\textbf{Fig. 5c}).

\begin{figure}[tbp]
\centering
\includegraphics[width=\linewidth,height=0.78\textheight,keepaspectratio]{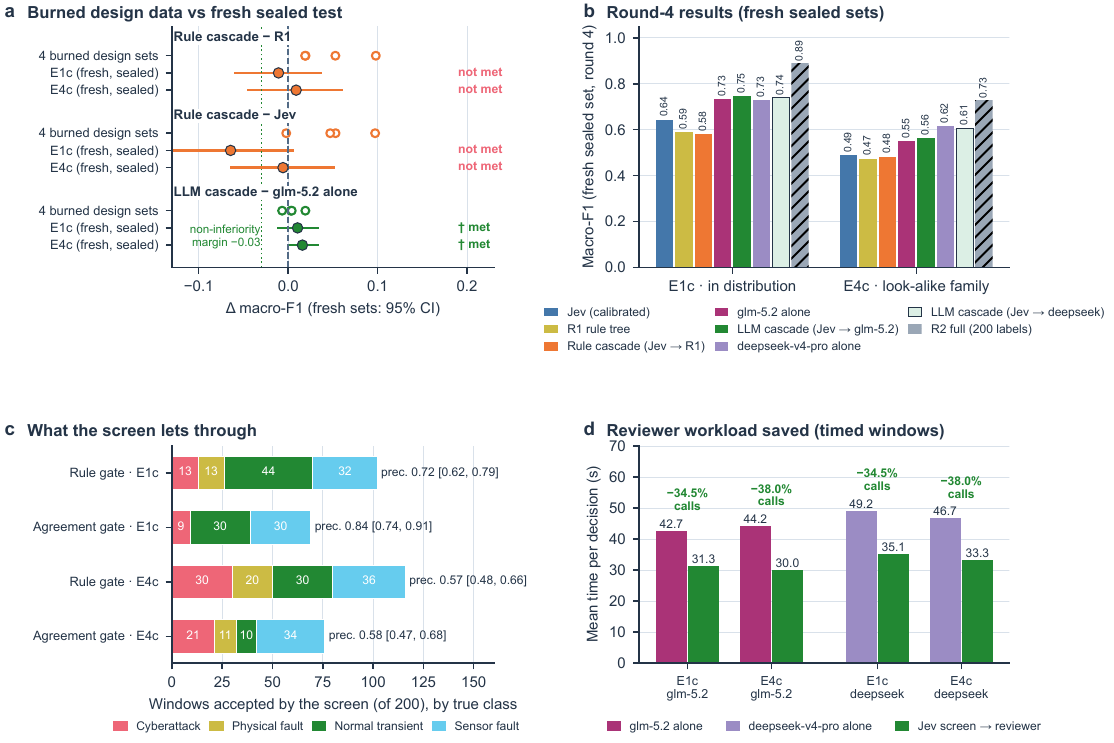}
\caption{Jev as a first-line screen. (a) Pre-registered cascade comparisons in macro-F1, on the four design and confirmation sets (E1r2, E2r2, E1r3, E3; light, n = 4) and on the two fresh sealed sets (E1c, E4c; dark, n = 2). The dotted line is the non-inferiority margin (\ensuremath{-}0.03) of the LLM cascade; ``met'' counts the fresh-set primary tests that rejected their null hypothesis after Holm correction over seven tests. (b) Accept rate and accept precision of the rule gate (every benign Jev verdict) and the agreement gate (benign Jev verdicts confirmed by R1) over the eight C-Town sealed sets. (c) Windows accepted per set by true class, for the same two gates. (d) Time per decision of each reviewer alone and behind the Jev screen, on the 40 round-4 windows timed one request at a time; decisions settled by the screen take only the Jev time. Bars show the mean, error bars \ensuremath{\pm}1 s.d. and white dots the individual sealed sets (a-c) or decisions (d); the mean is printed above each bar. With two fresh sets, the s.d. in (a) only indicates their spread.}
\label{fig:5}
\end{figure}

On the unseen family, accept precision was lower, 0.84 {[}0.74, 0.91{]}
on E1c against 0.58 {[}0.47, 0.68{]} on E4c, and 21 of the 76 accepted
E4c windows were attacks (12 masked duty swaps and 9 level-spoof
attacks) and 10 were pump trips (\textbf{Note S10}; \textbf{Fig. S6}).
The reviewers themselves recognised few of these attacks, so cyber
recall was almost unchanged by the screen, but the number of true
attacks that received any non-benign verdict fell from 20 to 15 of 50
with glm-5.2 and from 30 to 23 of 50 with deepseek-v4-pro (39 \ensuremath{\to} 37 and
38 \ensuremath{\to} 37 in distribution). Masked duty swaps were rarely recognised by
any method, fully supervised R2 included.

\subsection{Screen components}\label{screen-components}

Ablations isolate the working parts of the screen (\textbf{Fig. 6}). The
label-free prior correction (\textbf{Eqs. 1 and 2}) raised Jev from a
mean of 0.41 to 0.57 over rounds 1-3, because Jev's raw answers
over-weight physical faults; a transductive prior re-estimated on each
evaluated set was not better (0.58; \textbf{Fig. 6a}; \textbf{Note S4}).
Three physics checks computed from reported values alone (evidence
format 3) did not help Jev itself (0.60 against 0.63); the same evidence
did help rules, lifting R1 to 0.67 as R1b and to 0.68 when applied as an
override to the Jev verdict (\textbf{Fig. 6b}; \textbf{Note S6}). On a
new mechanism that mimics these attacks, pump wear, the override
assigned every window to the cyber class.

\begin{figure}[tbp]
\centering
\includegraphics[width=\linewidth,height=0.78\textheight,keepaspectratio]{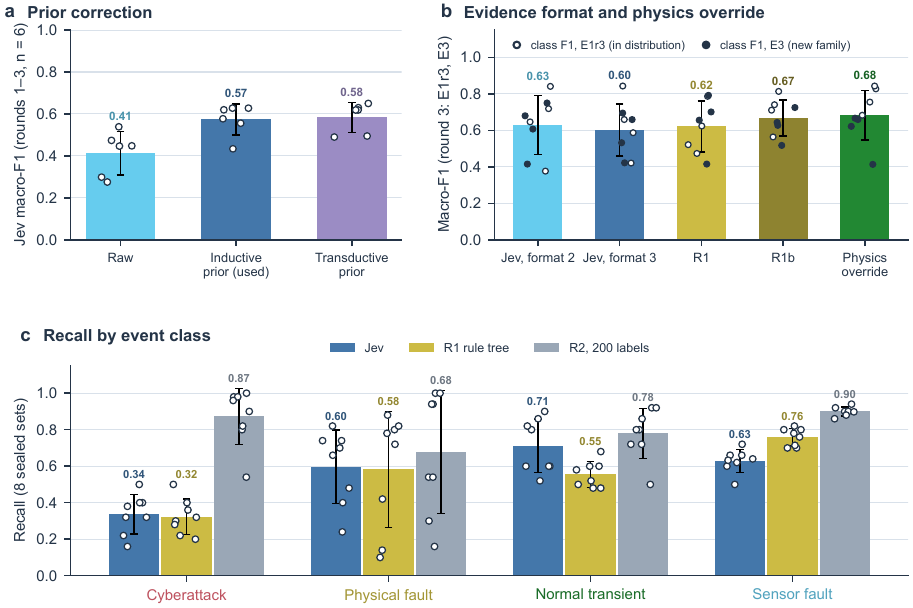}
\caption{Components of the Jev screen. (a) Macro-F1 of Jev with raw probabilities, with the inductive label-free prior used throughout the study, and with a transductive prior re-estimated on each evaluated set, over the six sealed sets of rounds 1-3. (b) Macro-F1 on the two round-3 sealed sets of Jev with evidence formats 2 and 3, of the rule tree R1, of R1 with a physics override (R1b), and of the same override applied to the Jev verdict. (c) Recall by true class of Jev, R1 and R2 trained on 200 labels over the eight C-Town sealed sets. Bars show the mean, error bars \ensuremath{\pm}1 s.d. and white dots the individual sealed sets; the mean is printed above each bar. With two sets in (b), the s.d. only indicates their spread.}
\label{fig:6}
\end{figure}

Concealed manipulation was the hardest class for the training-free
methods. Over the eight sealed sets its mean recall was 0.34 for
cyberattacks, against 0.60-0.71 for the other classes, and R1 was no
better (0.32), whereas supervised R2 recalled 0.87 (\textbf{Fig. 6c};
\textbf{Fig. S3}). The screen should therefore settle benign windows and
send suspected attacks onward. A preliminary study on the public BATADAL
data points the same way (\textbf{Note S8}; \textbf{Fig. S4}): asked to
separate 64 attack windows from 43 benign-alarm windows, Jev reached an
area under the receiver operating characteristic curve of 0.884,
statistically indistinguishable from Mahalanobis distance (0.934) and
below a supervised classifier (0.967), and used as a veto on a rule's
alarms it cut the false-positive rate from 25.6\% to 7.0\% at a
true-positive rate of 81.2\%.

\section{Discussion}\label{discussion}

\subsection{A fast first tier}\label{a-fast-first-tier}

A water utility does not need one classifier that is best at everything.
It needs a pipeline whose cheap tiers handle the common cases reliably
and whose expensive tiers are reserved for the contested ones, and our
results place a typed-judgment model in the first tier of that pipeline.
Without training data, Jev decides in about a second at the accuracy of
an expert-written rule tree. It keeps that accuracy on event subtypes
for which a supervised model has no labels, with a margin of 0.36-0.42
macro-F1 that replicated in four independent sealed rounds, it
outperforms supervised learning when fewer than about four labelled
events per class exist, and it decides 20-40 times faster than frontier
LLMs.

The fresh sealed test confirmed the gated cascade. The gated LLM cascade
kept its reviewer's accuracy while removing about a third of the
reviewer's workload, both in distribution and on an event family that no
design step had seen. Confirmation on fresh sealed sets is therefore an
integral part of designing such a cascade. The value of a fast screen
also lies less in its own accuracy than in the gate that decides which
of its verdicts to trust: agreement between two independent,
training-free judges on a benign verdict was a far better gate than
Jev's verdict alone.

Speed matters for the operating model of a utility, not only for
convenience. Hourly SCADA windows arrive continuously, and during a
multi-alarm incident dozens of windows may need triage within an hour.
At 30-50 s per decision, one request stream to a frontier LLM can review
about 80-120 windows per hour, and far fewer when the service is loaded,
whereas at about 1.4 s per decision Jev can screen about 2,500. A gated
screen that settles a third of the windows as benign removes that share
of the expensive review without lowering the reviewer's accuracy, and it
keeps the reviewer on every window in which either Jev or the rule tree
suspects an attack or an asset fault.

\subsection{Rival explanations}\label{rival-explanations}

Two rival explanations deserve attention. The subtype-level advantage
might show only that R2 is a weak learner. Against this, the same R2 is
excellent in distribution (0.887-0.934), and its loss of accuracy is
subtype-specific, complete on bursts and demand changes and absent on
sensor offsets, which is the signature of a mechanism the labels never
showed rather than of an under-powered model. The cascade result, in
turn, might reflect the reviewer rather than the screen. The screen
cannot add accuracy that the reviewer lacks; what it contributes is the
decision not to call the reviewer, and non-inferiority on both fresh
sets shows that this decision was safe on average.

\subsection{Applicability to other
networks}\label{applicability-to-other-networks}

The round-4 screen, calibration and cascades were applied without
refitting to EPANET Net3 and Net1 in two further sealed rounds
(\textbf{Note S11};~\textbf{Fig. S7}). The gated cascade stayed within
the non-inferiority margin of its reviewer on all four transfer sets
(\textbf{Fig. S8}; \textbf{Table S3}), and with the evaluated subtype
removed from the C-Town labels, Jev again exceeded the supervised model
on both transfer networks (\textbf{Fig. S11}).~All metrics of the
transfer sets are given in \textbf{Fig. S9}, their realised design,
reviewer time and gate statistics in \textbf{Fig. S10}, and the
label-scarcity curves in \textbf{Fig. S12.}~The screen, its calibration
and its gate therefore carry over to structurally different networks
without refitting.

\subsection{Limitations}\label{limitations}

This study has limitations. All windows were simulated, and one
hydraulic model generated both the events and the twin, so field data
would add model error. Cloud LLM weights and serving conditions can
change without notice, and the LLM times were measured from one client
at one point in time rather than as a guaranteed service level. The
benign gate accepts the attacks that both Jev and R1 judge benign, so in
deployment it can be tightened, for example by also requiring a low Jev
probability for the attack class, and paired with physics-based
interlocks that act regardless of attribution. Concealed manipulation
remains the hardest class for Jev, and the best general-purpose LLMs are
more accurate than it. The ungated rule cascade, which exceeded R1 on
three of four design sets with gains of up to +0.098, performed on a par
with its own rule tree on both fresh sets, so we do not claim that Jev
improves a rule tree; the gated cascade also had a lower accept
precision on the unseen family than in distribution. On Net1, a network
with one tank, one pump and four monitored junctions, pipe bursts
produced signatures close to those of demand changes and were rarely
recognised by Jev or the frontier reviewers; the agreement gate deferred
them to review because R1 did not confirm the benign verdict. Before a
utility lets a training-free screen settle windows on its own network,
it should therefore confirm on simulated events that every class of
event actually trips a check there.

\section{Conclusions}\label{conclusions}

This study tested whether a training-free typed-judgment model can serve
as the first tier of SCADA anomaly triage in WDNs. On a four-class
cause-attribution benchmark built on C-Town and evaluated in sealed,
pre-registered rounds, Jev needed neither labels nor rule engineering
and only a label-free prior correction. It attributed anomalies in about
1.2 s with the accuracy of an expert-written rule tree, and it decided
20-40 times faster than frontier LLMs. These properties suit it to a
first-pass screen rather than to a final verdict.

The main advantage of the screen lies where supervised learning is
weakest. On event subtypes absent from the training labels, Jev exceeded
a supervised classifier by 0.36-0.42 macro-F1 in all four C-Town rounds,
because a decision surface fitted to known subtypes did not extend to a
new mechanism. Jev also outperformed the classifier whenever fewer than
about four labelled events per class were available. It did not,
however, close the gap to a fully supervised model on genuinely new
event families, and its recall of concealed cyberattacks remained low.
The screen should therefore settle benign windows and pass suspected
attacks onward instead of dismissing them.

Used in this way, the screen reduced the load on deliberate review
without reducing its accuracy. Accepting a benign Jev verdict only when
the rule tree agreed spared an LLM reviewer 35-38\% of windows on fresh
sealed sets, with no loss of macro-F1, whereas accepting every benign
Jev verdict did not improve on its components; the agreement with an
independent judge was thus essential.

For utilities, the practical recommendation is a tiered triage: a fast,
training-free screen first, whose benign verdicts are accepted only when
an independent judge agrees, followed by rule-based or LLM review for
every window the screen does not settle, with physical interlocks
retained as the last line of defence. Validation on field SCADA records
and a stronger treatment of concealed manipulation are the next steps
before such a screen enters operation.

\section*{Declaration of competing interest}\label{declaration-of-competing-interest}

The authors declare no competing interests.

\section*{CRediT authorship contribution statement}\label{credit-authorship-contribution-statement}

\textbf{Tianwei Mu:} Conceptualization, Methodology, Software,
Investigation, Formal analysis, Visualization, Writing - original draft,
Funding acquisition. \textbf{Shengyan Jiang:} Data curation, Validation,
Writing - review \& editing. \textbf{Mingzhe Yuan:} Supervision,
Methodology, Writing - review \& editing. \textbf{Qing Luo:}
Supervision, Resources, Writing - review \& editing. \textbf{Min Xiao:}
Supervision, Project administration, Writing - review \& editing.
\textbf{Wenhong Wang:} Software, Validation. \textbf{Jun Li:} Writing -
review \& editing. \textbf{Manhong Huang:} Supervision, Writing - review
\& editing.

\section*{Code and Data availability}\label{code-and-data-availability}

C-Town is the public BATADAL network of (Taormina et al., 2018), and
Net3 and Net1 are distributed with EPANET and WNTR. The hydrojev Python
library, the scenario generators, the frozen questions, prompts and
design files, every Jev and LLM response, the analysis and figure
scripts, and the result files and source data behind every number and
figure in this article are available at
\url{https://github.com/mutianwei521/hydrojev}.

\section*{Acknowledgements}\label{acknowledgements}

This research was supported by the Guangdong Basic and Applied Basic
Research Foundation (No.~2026A1515011817), the Key Laboratory of
Eco-restoration of Regional Contaminated Environment (Shenyang
University), Ministry of Education (No.~KF-26-11), and the Liaoning
Provincial Education Department Fund Project ``Digital twin-driven
global resilience assessment model and dynamic simulation research''
(No.~202464252).

\section*{References}\label{references}

Abokifa, A.A., Haddad, K., Lo, C., Biswas, P., 2019. Real-time
identification of cyber-physical attacks on water distribution systems
via machine learning--based anomaly detection techniques. J. Water
Resour. Plan. Manag. 145 (1), 04018089.
\url{https://doi.org/10.1061/(asce)wr.1943-5452.0001023}

Alnegheimish, S., Nguyen, L., Berti-Equille, L., Veeramachaneni, K.,
2024. Can large language models be anomaly detectors for time series?,
in: 2024 IEEE 11th International Conference on Data Science and Advanced
Analytics (DSAA). IEEE, pp.~1--10.
\url{https://doi.org/10.1109/DSAA61799.2024.10722786}

Chandy, S.E., Rasekh, A., Barker, Z.A., Shafiee, M.E., 2019. Cyberattack
detection using deep generative models with variational inference. J.
Water Resour. Plan. Manag. 145 (2), 04018093.
\url{https://doi.org/10.1061/(asce)wr.1943-5452.0001007}

Chen, L., Zaharia, M., Zou, J., 2023. FrugalGPT: how to use large
language models while reducing cost and improving performance.
arXiv:2305.05176. \url{https://doi.org/10.48550/arXiv.2305.05176}

Eggimann, S., Mutzner, L., Wani, O., Schneider, M.Y., Spuhler, D., Moy
de Vitry, M., Beutler, P., Maurer, M., 2017. The potential of knowing
more: A review of data-driven urban water management. Environ. Sci.
Technol. 51 (5), 2538--2553.
\url{https://doi.org/10.1021/acs.est.6b04267}

Farquhar, S., Kossen, J., Kuhn, L., Gal, Y., 2024. Detecting
hallucinations in large language models using semantic entropy. Nature
630 (8017), 625--630. \url{https://doi.org/10.1038/s41586-024-07421-0}

Fu, G., Jin, Y., Sun, S., Yuan, Z., Butler, D., 2022. The role of deep
learning in urban water management: A critical review. Water Res. 223,
118973. \url{https://doi.org/10.1016/j.watres.2022.118973}

Geifman, Y., El-Yaniv, R., 2017. Selective classification for deep
neural networks, in: Advances in Neural Information Processing Systems
30 (NeurIPS 2017). pp.~4878--4887.
\url{https://arxiv.org/abs/1705.08500}

Goldstein, Y., Perelman, G., Schuster, A., Ostfeld, A., 2026. Large
language models for water distribution systems modeling and
decision-making. Environ. Model. Softw. 205, 107148.
\url{https://doi.org/10.1016/j.envsoft.2026.107148}

Hassanzadeh, A., Rasekh, A., Galelli, S., Aghashahi, M., Taormina, R.,
Ostfeld, A., Banks, M.K., 2020. A review of cybersecurity incidents in
the water sector. J. Environ. Eng. 146 (5), 03120003.
\url{https://doi.org/10.1061/(asce)ee.1943-7870.0001686}

Holm, S., 1979. A simple sequentially rejective multiple test procedure.
Scand. J. Stat. 6 (2), 65--70.
\url{https://www.jstor.org/stable/4615733}

Housh, M., Ohar, Z., 2018. Model-based approach for cyber-physical
attack detection in water distribution systems. Water Res. 139,
132--143. \url{https://doi.org/10.1016/j.watres.2018.03.039}

Ji, Z., Lee, N., Frieske, R., Yu, T., Su, D., Xu, Y., Ishii, E., Bang,
Y.J., Madotto, A., Fung, P., 2023. Survey of hallucination in natural
language generation. ACM Comput. Surv. 55 (12), 248.
\url{https://doi.org/10.1145/3571730}

Kahneman, D., 2011. Thinking, Fast and Slow. Farrar, Straus and Giroux,
New York.

Klise, K.A., Bynum, M., Moriarty, D., Murray, R., 2017. A software
framework for assessing the resilience of drinking water systems to
disasters with an example earthquake case study. Environ. Model. Softw.
95, 420--431. \url{https://doi.org/10.1016/j.envsoft.2017.06.022}

Larsen, T.A., Hoffmann, S., L\"uthi, C., Truffer, B., Maurer, M., 2016.
Emerging solutions to the water challenges of an urbanizing world.
Science 352 (6288), 928--933.
\url{https://doi.org/10.1126/science.aad8641}

Rasekh, A., Hassanzadeh, A., Mulchandani, S., Modi, S., Banks, M.K.,
2016. Smart water networks and cyber security. J. Water Resour. Plan.
Manag. 142 (7), 01816004.
\url{https://doi.org/10.1061/(asce)wr.1943-5452.0000646}

Richards, C.E., Tzachor, A., Avin, S., Fenner, R., 2023. Rewards, risks
and responsible deployment of artificial intelligence in water systems.
Nat. Water 1 (5), 422--432.
\url{https://doi.org/10.1038/s44221-023-00069-6}

Rossman, L.A., 2000. EPANET 2 Users Manual. Report EPA/600/R-00/057.
U.S. Environmental Protection Agency, Cincinnati, OH.

Saerens, M., Latinne, P., Decaestecker, C., 2002. Adjusting the outputs
of a classifier to new a priori probabilities: a simple procedure.
Neural Comput. 14 (1), 21--41.
\url{https://doi.org/10.1162/089976602753284446}

Taormina, R., Galelli, S., 2018. Deep-learning approach to the detection
and localization of cyber-physical attacks on water distribution
systems. J. Water Resour. Plan. Manag. 144 (10), 04018065.
\url{https://doi.org/10.1061/(asce)wr.1943-5452.0000983}

Taormina, R., Galelli, S., Tippenhauer, N.O., Salomons, E., Ostfeld, A.,
2017. Characterizing cyber-physical attacks on water distribution
systems. J. Water Resour. Plan. Manag. 143 (5), 04017009.
\url{https://doi.org/10.1061/(asce)wr.1943-5452.0000749}

Taormina, R., Galelli, S., Tippenhauer, N.O., Salomons, E., Ostfeld, A.,
Eliades, D.G., Aghashahi, M., Sundararajan, R., Pourahmadi, M., Banks,
M.K., Brentan, B.M., Campbell, E., Lima, G., Manzi, D., Ayala-Cabrera,
D., Herrera, M., Montalvo, I., Izquierdo, J., Luvizotto, E., Chandy,
S.E., Rasekh, A., Barker, Z.A., Campbell, B., Shafiee, M.E., Giacomoni,
M., Gatsis, N., Taha, A., Abokifa, A.A., Haddad, K., Lo, C.S., Biswas,
P., Pasha, M.F.K., Kc, B., Somasundaram, S.L., Housh, M., Ohar, Z.,
2018. Battle of the attack detection algorithms: disclosing cyber
attacks on water distribution networks. J. Water Resour. Plan. Manag.
144 (8), 04018048.
\url{https://doi.org/10.1061/(asce)wr.1943-5452.0000969}

TypeSafe AI, n.d. System One. TypeSafe AI documentation.
\url{https://docs.typesafe.ai/concepts/system-one} (accessed 23
September 2026).

Viola, P., Jones, M., 2001. Rapid object detection using a boosted
cascade of simple features, in: Proceedings of the 2001 IEEE Computer
Society Conference on Computer Vision and Pattern Recognition (CVPR
2001). IEEE, pp.~I-511--I-518.
\url{https://doi.org/10.1109/cvpr.2001.990517}

Wang, J., Fu, G., Savic, D., 2026. Leveraging large language models for
automating water distribution network optimization. Water Res. 288,
124536. \url{https://doi.org/10.1016/j.watres.2025.124536}

Wilson, E.B., 1927. Probable inference, the law of succession, and
statistical inference. J. Am. Stat. Assoc. 22 (158), 209--212.
\url{https://doi.org/10.1080/01621459.1927.10502953}

Zhou, H., Wan, X., Proleev, L., Mincu, D., Chen, J., Heller, K., Roy,
S., 2024. Batch calibration: rethinking calibration for in-context
learning and prompt engineering, in: The Twelfth International
Conference on Learning Representations (ICLR 2024).
\url{https://arxiv.org/abs/2309.17249}
\clearpage
\setcounter{figure}{0}
\setcounter{table}{0}
\renewcommand{\thefigure}{S\arabic{figure}}
\renewcommand{\thetable}{S\arabic{table}}
\renewcommand{\theHfigure}{S\arabic{figure}}
\renewcommand{\theHtable}{S\arabic{table}}
\section*{Supplementary Information}

\section*{Note S1. Benchmark generation and splits}\label{note-s1.-benchmark-generation-and-splits}

Every scenario was simulated with the EPANET engine of WNTR on the
unmodified C-Town model at hourly steps over 48-54 h (\textbf{Fig.
S1a}); \textbf{Fig. S1b} shows the SCADA channels read by the evidence
state. Demand multipliers carry 4\% independent multiplicative noise.
The reported historian adds Gaussian measurement noise to every channel
and then applies the telemetry manipulation or instrument fault of the
event. Onsets are drawn uniformly on 24-48 h, and the analysed window
ends 2-6 h after onset. A scenario is regenerated from the next seed of
its deterministic ladder when the event is not yet physically expressed
in the window, for example when the tank that an attack over-pumps has
not reached its stop set-point, or when a stuck sensor reads within
noise of the true value.

Each split has its own seed base, and the ladders of different splits
are disjoint. \textbf{Table S1} lists every split of the six rounds with
its seed base, size, subtype family and use.

Two deviations from the protocol occurred and are disclosed. In round 1
the regeneration ladder, which adds 100,000 to the seed at each attempt,
made 3 of the 40 E1 windows identical to development windows and 1
development window identical to a training window. The round-1 results
were kept unchanged, which gives them a slight optimistic bias, and
every later split uses seed bases at or above 5,050,000. In round 3 one
Jev call on an evidence-format-3 window failed; by the rule that no
request is re-sent without confirming that its record is absent, it was
not repeated, and the remaining 199 E1r3 windows were scored.

\section*{Note S2. Evidence state, screen question and
baselines}\label{note-s2.-evidence-state-screen-question-and-baselines}

\textbf{Evidence state.} Each state is a JSON document in six parts: a
neutral description of the network and its SCADA channels; the control
configuration (level set-points and timed rules); the list of
consistency checks, each with its status (normal or violated),
statistic, threshold and base rate in 80 event-free runs; the ten
channels with the largest twin residuals as z-scores; a signed summary
(direction and z-score of the system-demand residual, whether the
pressure deviation is widespread or localized and its sign, the signs of
the tank residuals and the number of channels that corroborate the
largest anomaly); and the source-meter balance, which compares the
source outflow meters with the station pump flows.

Evidence format 2 carries ten checks. Three compare pumps with their own
measurements or with the twin: pump status against pump flow, pump head
gain against the twin, and pump flow against the twin. Two concern the
tanks: the level change against the tank inflow meter (mass balance) and
the level against the twin. Two compare network-wide quantities with the
twin, the pressures and the system demand. Two apply the control logic
to the reported actuator states, once with the reported tank levels and
once with the twin's tank levels. The last is the source-meter balance.
Thresholds are 97.5\% quantiles over 80 event-free runs with independent
seeds, and 30 for the balance.

Before any call, the whole state is scanned for a fixed list of
forbidden tokens: the class names; mechanism words such as cyber,
attack, spoof, replay, burst, leak, stuck, drift, surge, overpump,
masked, closure and gain error; label-carrying field names such as
scenario, subtype and label; and the network names. A state with any hit
is not written, and no hit occurred in any round.

\textbf{Screen question.} The question was fixed on the 40 development
windows of round 1 before any evaluation and was used unchanged in every
round. It is a single-choice question over five options. Its
instructions read: ``Using only the supplied evidence for this six-hour
SCADA window, decide which cause best explains it. Treat the four causes
as equally likely before looking at the evidence. The twin is a model
with nominal consumption and nominal assets, not a measurement: a
measurement that disagrees with the twin shows that the world differs
from the model, not that the measurement is false. Only a disagreement
between measurements, or between a reported actuator state and the
measurements it should produce, is a contradiction. Reason in this
order. First, is there such a contradiction (for example a violated
check that compares two reported quantities with each other)? If yes,
decide between concealed manipulation and a faulty instrument by whether
one channel alone is wrong while every channel related to it is normal.
Second, if there is none, decide how the world differs from the model:
consumption changed network-wide (pressures, tank levels, source outflow
and pump flows all shifted in the direction that more or less
consumption implies), an operator switched an actuator, or an asset
fault acts at one location. Give every cause a non-zero probability that
reflects how well the evidence fits its definition.''

The five options are defined verbatim as follows. Cyberattack:
``concealed manipulation of control or telemetry: a reported actuator
status or reading is contradicted by independent, physically related
measurements (meters of other assets, pressures, flows, mass balance),
so the reported operation does not match what the network is physically
doing. Concealment may make several reported channels agree with one
another while the physics measured elsewhere disagrees; a single faulty
instrument cannot explain the pattern.'' Physical fault: ``a real fault
of a network asset (pipe, fitting, pump or valve) that the twin does not
model: measurements deviate from the twin, yet independent meters agree
with one another and with the reported actuator states. The effect is
concentrated around one location or asset rather than acting on the
whole network in the same way.'' Normal transient: ``ordinary, honestly
reported operation: consumption higher, lower or earlier/later than
nominal acting on the whole network coherently, or an actuator switched
by an operator. The twin assumes nominal consumption and only knows the
configured control logic, so under this cause its residuals, its
estimated tank levels and the control-logic checks can all disagree with
operation, while the independent measurements agree with one another and
confirm every reported actuator state.'' Sensor fault: ``a single
measuring instrument misreports: one channel deviates from the twin and
from the channels physically related to it, while those related
channels, the actuators and the hydraulics show nothing unusual.''
Insufficient evidence: to be used ``only when the evidence is missing or
genuinely contradictory so that none of the other four explanations can
be preferred.''

\textbf{Rule tree R1.} R1 was frozen before the first Jev call. Its
rules are evaluated in order, and the first rule that applies decides
(\textbf{Table S2}). R1b, tested in round 3, prepends a physics override
built on the three checks of evidence format 3 (\textbf{Note S6}): it
returns cyberattack whenever a stopped station has a discharge head
inconsistent with its tank, a pump operates off its curve, or the
integrated inflow crosses the stop set-point without a stop.

\textbf{Supervised classifier R2.} R2 is a standard scaler followed by
multinomial logistic regression (C = 1, balanced class weights, at most
5,000 iterations). Its features are the numerical content of the state:
every check statistic divided by its threshold, the signed summary
quantities and the count of anomalous channels, so R2 sees exactly what
Jev sees. R2 full is trained on the 200 C-Town training windows, R2 LOSO
on the same windows without the evaluated subtype. In rounds 5 and 6, R2
transfer is R2 full applied to the target network, and R2 own network is
refitted on 200 labelled windows of the target network, as a reference
that needs labelled data from that network.

\section*{Note S3. LLM protocol}\label{note-s3.-llm-protocol}

Each LLM received a system message restating the task, followed by a
user message that contained the screen question and definitions
(\textbf{Note S2}) and the same state that Jev received. Sampling was
fixed at temperature 0 with seed 20260925, the output was constrained to
a JSON schema with five numeric fields in {[}0, 1{]}, and streaming was
off. Probabilities were renormalized; an answer that could not be parsed
or had zero total mass was counted as invalid and wrong. A timed-out
request was retried only after its record had been confirmed absent, and
no answer was re-sampled for its content. The model name resolved by the
server was recorded with every response. Each model was calibrated like
Jev, with its own mean answer on the unlabelled development windows
(\textbf{Eqs. 1 and 2}).

Seven models answered every window of the two round-2 sealed sets
validly. On E1r2 and E2r2 respectively, deepseek-v4-pro scored a
macro-F1 of 0.722 and 0.637, glm-5.2 0.706 and 0.697, gemma4 0.641 and
0.470, gemma4:31b 0.631 and 0.475, gpt-oss:120b 0.615 and 0.568,
minimax-m2.7 0.606 and 0.478, and gpt-oss:20b 0.551 and 0.532, against
0.628 and 0.434 for Jev (\textbf{Fig. 4a}). The frontier models produced
a median of about 5,600-8,600 (deepseek-v4-pro) and 2,300-2,900
(glm-5.2) output tokens per decision, gpt-oss:120b about 1,000 and
gemma4 72, whereas Jev returns a typed answer without generated text. Of
the 14 primary two-sided tests (\textbf{Table S3}), four were
significant after Holm correction, all in favour of the LLM:
deepseek-v4-pro on E1r2 (\ensuremath{-}0.094 {[}\ensuremath{-}0.157, \ensuremath{-}0.033{]}) and on E2r2
(\ensuremath{-}0.203 {[}\ensuremath{-}0.278, \ensuremath{-}0.131{]}), glm-5.2 on E2r2 (\ensuremath{-}0.264 {[}\ensuremath{-}0.337,
\ensuremath{-}0.191{]}) and gpt-oss:120b on E2r2 (\ensuremath{-}0.134 {[}\ensuremath{-}0.208, \ensuremath{-}0.060{]}).

\section*{Note S4. Prior
calibration}\label{note-s4.-prior-calibration}

Three calibrations of Jev were compared on the six sealed sets of rounds
1-3: the argmax of the raw probabilities (\textbf{Eq. 2} with
\(\pi_{c} = 1\)), the frozen inductive calibration that divides by the
development prior (\textbf{Eqs. 1 and 2}), and a transductive
leave-one-out prior re-estimated without labels on the other windows of
the evaluated set,

\begin{equation}
\pi_{c}^{( - w)} = \max\left( 0.01,\ \frac{1}{|\mathcal{E}|-1}\sum_{v\in\mathcal{E},\, v 
\neq w}^{}p_{c}(v) \right)
\tag{S1}
\end{equation}

where \(\mathcal{E}\) is the evaluated sealed set and
\(\pi_{c}^{( - w)}\) replaces \(\pi_{c}\) in \textbf{Eq. 2} when window
\(w\) is scored. \textbf{Eq. S1} uses no labels but does use the other
windows of the evaluated set, so it was reported as an ablation and
never pre-registered as a primary method.

The development prior was \ensuremath{\pi} = (0.302, 0.429, 0.146, 0.121) for
cyberattack, physical fault, normal transient and sensor fault. The
frozen calibration raised macro-F1 on every set: from 0.447 to 0.620 on
E1, 0.299 to 0.556 on E2, 0.539 to 0.628 on E1r2, 0.276 to 0.434 on
E2r2, 0.474 to 0.628 on E1r3 and 0.448 to 0.577 on E3. Raw Jev
over-predicts physical faults; on E1r2 it recalled 96\% of bursts but
only 32\% of demand falls (\textbf{Fig. S2a}), and dividing by the
development prior removes this bias. The transductive prior of
\textbf{Eq. S1} changed macro-F1 by \ensuremath{-}0.061 (E2) to +0.056 (E2r2)
relative to the frozen calibration, with no consistent gain.

\section*{Note S5. Recall by
subtype}\label{note-s5.-recall-by-subtype}

The recall by subtype on the sealed sets of rounds 2 and 3 is shown in
\textbf{Fig. S2}. On the in-distribution sets E1r2 and E1r3
(\textbf{Fig. S2a,b}), R2 LOSO recalled none of the bursts, demand rises
or masked overpump attacks and at most 0.05 of the pump-off attacks
spoofed as running, whereas calibrated Jev recalled 0.74-0.82 of the
bursts and 0.88-0.95 of the demand rises. Jev's weakest subtypes were
the concealed attacks: it recalled 0.08 and 0.25 of the masked overpump
attacks and 0.34-0.40 of the pump-off attacks. On the new families,
whose subtypes are all unseen by R2 (\textbf{Fig. S2c,d}), the hardest
subtype for every method was the pipe closure of E2r2 (Jev 0.24, R1
0.14, R2 full 0.16), and R2 full recalled none of the pump-wear windows
of E3, which Jev recalled at 0.65 and R1 at 0.83.

The confusion matrices of calibrated Jev on all eight C-Town sealed sets
are given in \textbf{Fig. S3}. On the six sets of rounds 1-3 its recall
ranged over 0.16-0.50 for cyberattacks, 0.24-0.82 for physical faults,
0.52-0.90 for normal transients and 0.50-0.66 for sensor faults. The
cyberattack class was its weakest on five of these six sets, the
exception being the round-1 new-family set E2, where physical faults
were recalled at 0.40 and cyberattacks at 0.50. The round-4 counterparts
are \textbf{Figs. S5 and S6} (\textbf{Note S10}), and those of rounds 5
and 6 are in \textbf{Fig. S11} (\textbf{Note S11}).

\section*{Note S6. Evidence format 3 and physics
override}\label{note-s6.-evidence-format-3-and-physics-override}

Evidence format 3 adds three checks computed only from reported values.
The first compares the head gain predicted by the pump curve at the
reported flow with the reported discharge-minus-suction head. The second
requires that, when every pump of a station is reported off, the
discharge head equals the downstream tank head. The third compares the
tank level integrated from the inflow meter over the window with the
level at which the control logic should have stopped the pumps.
Thresholds come from 80 event-free runs, and the format-2 and format-3
states of a window come from the same simulation.

On the round-3 sealed sets, format 3 lowered Jev's macro-F1 slightly and
not significantly: 0.628 against 0.643 in format 2 on E1r3 (\ensuremath{\Delta} \ensuremath{-}0.018
{[}\ensuremath{-}0.061, 0.025{]}) and 0.577 against 0.613 on E3 (\ensuremath{\Delta} \ensuremath{-}0.036 {[}\ensuremath{-}0.085,
0.013{]}). The physics override helped where the attacks it targets
occur: on E1r3 it raised R1 from 0.602 to 0.707 (R1b) and Jev to 0.777
(override applied to the Jev verdict), and it raised the recall of
pump-off attacks spoofed as running from 0.34 to 0.97. On a mechanism
that mimics them the override changed the verdict: on E3, pump-wear
recall fell from 0.65 for Jev to 0.00 with the override, every such
window being read as an attack, and the hybrids scored 0.629 (R1b) and
0.591 (Jev with override) against 0.641 for R1 (\textbf{Table S3}).

\section*{Note S7. Label scarcity}\label{note-s7.-label-scarcity}

R2 was retrained on \(k\) labelled training windows per class (\(k\) =
1, 2, 4, 8, 16, 40), drawn at random without replacement, with 200 draws
per \(k\); each draw was scored on the sealed set, and the crossover
\(k^{*}\) was computed with \textbf{Eq. 8}. On seven of the eight C-Town
sealed sets \(k^{*}\) was 4 (\textbf{Fig. 3}): at \(k\) = 2 the mean R2
macro-F1 was still below Jev's, and R2 fell below Jev in 53-88\% of the
draws, depending on the set. The exception was the round-1 new-family
set E2 (40 windows), where the mean R2 macro-F1 reached Jev's 0.556 only
at \(k\) = 40. Against the rule cascade of round 4, \(k^{*}\) was 2 on
E1c and 4 on E4c. On the transfer networks, where R2 was refitted on
labels of the evaluated network, \(k^{*}\) was 8 and 4 on Net3 and 1 and
4 on Net1 (\textbf{Fig. S12}; \textbf{Note S11}).

\section*{Note S8. Benign-alarm veto on
BATADAL}\label{note-s8.-benign-alarm-veto-on-batadal}

A separate study asked a narrower question: can Jev tell attack windows
from benign alarm windows in the public BATADAL data (Taormina et al.,
2018)? It used 107 windows, 64 of them attacks from 15 segments, all
flagged by at least one prior detector. BATADAL is simulated on the same
C-Town network, and its 12 pressure sensors are marked in \textbf{Fig.
S1b}.

Jev reached a ROC AUC of 0.884 (\textbf{Fig. S4a}) and was significantly
better than six of the seven prior detectors: the PCA residual (0.787),
the CPDZ residual (0.629), the GEpFM drift detector (0.566), the RAT
covariance detector (0.514), the autoencoder (0.495) and the Isolation
Forest (0.390). It was statistically indistinguishable from the
Mahalanobis detector (0.934; \ensuremath{\Delta} \ensuremath{-}0.050 {[}\ensuremath{-}0.110, 0.009{]}) and worse
than R2 trained on development data (0.967; \ensuremath{\Delta} \ensuremath{-}0.082 {[}\ensuremath{-}0.144,
\ensuremath{-}0.030{]}). In a veto cascade in which Jev must confirm each R1 alarm,
the false-positive rate of R1 fell from 25.6\% to 7.0\% at a
true-positive rate of 81.2\% (\textbf{Fig. S4b}), with a balanced
accuracy of 0.871 against 0.856 for R1 and 0.914 for R2. On the 73
windows that R1 flagged, Jev's AUC was 0.843 against 0.808 for
Mahalanobis. Jev can therefore act as a filter over another method's
decisions.

\section*{Note S9. Time to decision}\label{note-s9.-time-to-decision}

All 2,722 Jev calls made before round 4 were analysed. The
service-reported compute time had a median of 1.215 s (10th percentile
1.151 s, 90th percentile 1.458 s, mean 1.287 s, maximum 5.72 s), and the
client wall time a median of 1.463 s (90th percentile 1.706 s). The 400
round-4 calls gave a median compute time of 1.18 s and a median wall
time of 1.43 s, and rounds 5 and 6 gave the same medians within 0.03 s
(\textbf{Fig. S10c}).

Single training-window LLM requests sent without concurrency took 1.4 s
for gemma4, 7.9 s for gpt-oss:120b, 47 s for gpt-oss:20b and 56 s for
gemma4:31b, and longer for the frontier models: 82 s for glm-5.2, 97 s
for deepseek-v4-pro and 119 s for minimax-m2.7. Under the concurrent
load of round 2 the median time per decision was 600-1,600 s, and in the
round-3 confirmation run 382 s for glm-5.2 and 196 s for
deepseek-v4-pro. Sent one request at a time in round 4, the median was
30.6-31.7 s for glm-5.2 and 39.3-46.4 s for deepseek-v4-pro, 20-40 times
the Jev time. LLM time is dominated by generation (\textbf{Note S3}). R1
and R2 decide in well under 1 ms on a single CPU core, so the cascade
time of \textbf{Eq. 9} neglects R1.

\section*{Note S10. Cascades and round-4
results}\label{note-s10.-cascades-and-round-4-results}

\textbf{Design and confirmation.} The cascades were designed on the
round-2 sets, where Jev, R1 and all seven LLMs had answered every
window. The gate accepts only benign Jev verdicts because these were
Jev's most precise verdicts, and the LLM cascade additionally requires
agreement with R1 (\textbf{Eq. 4}), which keeps the reviewer on every
contested window at a modest cost in acceptance. On E1r2 and E2r2 the
rule cascade (\textbf{Eq. 3}) scored 0.626 and 0.531, against 0.628 and
0.434 for Jev and 0.606 and 0.478 for R1; the LLM cascade with glm-5.2
scored 0.699 and 0.717 against 0.706 and 0.697 for glm-5.2 alone, at
accept rates of 32\% and 21\% and accept precisions of 0.83 and 0.91;
with deepseek-v4-pro it scored 0.713 against 0.722 on E1r2 and 0.660
against 0.637 on E2r2.

The confirmation criterion was written before any round-3 LLM call. On
both round-3 format-2 sets it required the glm-5.2 cascade to reach at
least the macro-F1 of glm-5.2 alone minus 0.03, and an accept rate of at
least 0.15. Both reviewers answered all 800 round-3 requests validly,
and the criterion was met on both sets: the glm-5.2 cascade scored 0.713
against 0.710 alone on E1r3 and 0.736 against 0.732 on E3, at accept
rates of 30.5\% and 33.5\% and accept precisions of 0.80 and 0.84; the
deepseek-v4-pro cascade scored 0.707 against 0.690 and 0.719 against
0.717. The rule cascade scored 0.696 and 0.660 on these sets, above both
Jev (0.643 and 0.613) and R1 (0.598 and 0.641) (\textbf{Fig. 5a}). The
design was then frozen, together with the Jev and reviewer priors, the
seven primary hypotheses and the non-inferiority margin. Reviewer times
in the round-3 run, recorded with about 7-12 requests in flight, had
medians of 420-427 s for glm-5.2 and 378-509 s for deepseek-v4-pro;
because they overstate single-request latency, round 4 timed 20 windows
per set sent one request at a time.

\textbf{Round-4 sealed results.} Jev answered all 400 calls, and glm-5.2
and deepseek-v4-pro each returned 400 valid answers, so the
pre-specified fallback to the screen verdict was never used. The glm-5.2
cascade scored 0.745 on E1c and 0.565 on E4c, against 0.735 and 0.549
for glm-5.2 alone (\ensuremath{\Delta} +0.011 {[}\ensuremath{-}0.011, 0.033{]} and +0.016 {[}0.003,
0.033{]}; \textbf{Table S3}). The agreement gate settled 69 of the 200
E1c windows at an accept precision of 0.841 {[}0.737, 0.909{]} and 76 of
the 200 E4c windows at 0.579 {[}0.467, 0.683{]}, with Wilson intervals

\begin{equation}
\frac{\widehat{q} + \frac{z^{2}}{2n} \pm z\sqrt{\frac{\widehat{q}(1 - \widehat{q})}{n} + \frac{z^{2}}{4n^{2}}}}{1 + \frac{z^{2}}{n}}
\tag{S2}
\end{equation}

where \(\widehat{q}\) is the accept precision, \(n\) the number of
accepted windows and \(z\) = 1.96; \textbf{Eq. S2} is used for every
accept precision in this document. The ungated rule cascade met none of
its four superiority hypotheses: it scored 0.579 and 0.483, against
0.590 and 0.474 for R1 and 0.644 and 0.489 for Jev, and its rule gate
accepted 102 and 116 windows at precisions of 0.716 and 0.569. Jev beat
R2 LOSO by +0.417 {[}0.344, 0.488{]} on E1c. Neither Jev nor either
cascade reached R2 full, which scored 0.887 and 0.726 (\textbf{Fig.
S6}).

The wrong accepts of the agreement gate were few on E1c and concentrated
in two subtypes: 9 masked overpump attacks and 2 operator switches among
its 69 accepted windows. On E4c, 32 of the 76 accepted windows were
wrong: 12 masked duty swaps, 10 pump trips, 9 level-spoof-starve attacks
and 1 growing leak. Among the 50 true attacks of each set, the number
receiving any non-benign verdict was 39 for glm-5.2 alone and 37 for its
cascade on E1c, and 20 and 15 on E4c; for deepseek-v4-pro it was 38 and
37 on E1c and 30 and 23 on E4c. Jev gave a non-benign verdict to 37 and
20 of these attacks, R1 to 33 and 22, and R2 full to 48 and 29. On the
timed windows, the mean time per decision fell from 42.7 to 31.3 s (E1c)
and from 44.2 to 30.0 s (E4c) with glm-5.2, and from 49.2 to 35.1 s and
from 46.7 to 33.3 s with deepseek-v4-pro. The recall by subtype
(\textbf{Fig. S5}) shows that the masked overpump attack of E1c and the
masked duty swap of E4c were recognised by few training-free methods,
and that no method, R2 full included, recalled more than 0.38 of the
pump trips of E4c.

\section*{Note S11. Transfer to EPANET Net3 and Net1 (rounds 5 and
6)}\label{note-s11.-transfer-to-epanet-net3-and-net1-rounds-5-and-6}

\textbf{Design.} Net3 and Net1 (\textbf{Fig. S7}) were chosen because
they differ from C-Town in size, number of tanks and control structure.
Nothing fitted or chosen on C-Town was changed; only network properties
(channel set, residual and check thresholds, station balance
calibration) were recomputed with the frozen code on event-free runs.
Net1 was used with every demand pattern resampled to a 1-h step by
repeating each multiplier, which matches the hourly time base of the
event generator and leaves its event-free hydraulics unchanged. Before
each freeze, the only windows generated on the target network were a
realisability probe, a gate diagnostic of 30 bursts (Net3), design-check
seeds evaluated with the gates and R1 only (Net1) and the training split
for R2 own network, none of which had a Jev or LLM answer (\textbf{Table
S1}). Round 6 was frozen after round 5 had been analysed and carried its
design over unchanged. Each sealed set has 50 windows per class, and the
seed ladder redraws a window when the network cannot host the subtype or
the event trips no check (\textbf{Fig. S10a,b}). Per set, two primary
tests were pre-registered, non-inferiority of the glm-5.2 cascade to
glm-5.2 alone and superiority of Jev over R2 transfer, and they were
Holm-corrected within their round (\textbf{Eq. 7}). Round 5 retains its
pre-registered family of eight tests, which also covered four tests on
Net1 sets that were excluded from analysis, so its correction is
conservative.

\textbf{Round 5, Net3.} Jev answered all 400 calls, and both reviewers
returned 400 valid answers. On Net3 the realised subtype mix was close
to that of C-Town, with at most 52 redraws per class for non-realisable
subtypes and at most 22 for events that tripped no check. The glm-5.2
cascade was non-inferior to glm-5.2 alone on N3-Ref (\ensuremath{\Delta} \ensuremath{-}0.004 {[}\ensuremath{-}0.013,
0.001{]}). On N3-C the whole 95\% interval lay inside the margin (\ensuremath{-}0.007
{[}\ensuremath{-}0.026, 0.008{]}), but the p-value of 0.0117 missed the Holm
threshold of 0.0083 of the eight-test family, and the test is reported
as not met (\textbf{Table S3}). Jev scored 0.710 on N3-Ref, higher than
on C-Town, and tied R1 (\ensuremath{\Delta} \ensuremath{-}0.024 {[}\ensuremath{-}0.106, 0.058{]}), but R2 transfer
was more accurate than Jev on both Net3 sets (0.839 and 0.653 against
0.710 and 0.541; \textbf{Figs. S8 and S9}). The agreement gate settled
72 and 69 windows at accept precisions of 0.875 {[}0.779, 0.933{]} and
0.652 {[}0.534, 0.754{]}, and accepted 4 and 3 true attacks, against 17
for the rule gate on each set (\textbf{Fig. S8c}). With the evaluated
subtype removed from the C-Town labels, Jev beat R2 by +0.446 {[}0.378,
0.511{]} on N3-Ref (\textbf{Fig. S13b}).

\textbf{Round 6, Net1.} On Net1 a burst tripped a check in 50 of 161
draws, the benign transients of N1-Ref were a mix of demand falls (19),
demand rises (21) and operator switches (10), and the benign transients
of N1-C needed 45 redraws (\textbf{Fig. S10a,b}). Jev answered all 400
calls, and both reviewers returned 400 valid answers.

Three of the four round-6 primary tests were met (\textbf{Table S3}).
The glm-5.2 cascade was non-inferior to glm-5.2 alone on both sets
(\ensuremath{-}0.002 {[}\ensuremath{-}0.016, 0.012{]} and \ensuremath{-}0.005 {[}\ensuremath{-}0.024, 0.009{]}). Jev was
superior to R2 transfer on N1-C (+0.150 {[}0.086, 0.213{]}) but not on
N1-Ref, where R2 transfer was more accurate (\ensuremath{-}0.165 {[}\ensuremath{-}0.249,
\ensuremath{-}0.084{]}). On N1-Ref, R1 scored 0.668 and R2 transfer 0.602, whereas on
N1-Ref and N1-C Jev scored 0.437 and 0.466, glm-5.2 0.512 and 0.486, and
deepseek-v4-pro 0.511 and 0.493 (\textbf{Figs. S8a and S9}). The
agreement gate settled 52 and 62 windows at accept precisions of 0.808
{[}0.681, 0.892{]} and 0.694 {[}0.570, 0.794{]}, accepting 3 and 7 true
attacks and 3 and 8 physical faults. The rule gate accepted 154 and 133
windows at precisions of 0.409 and 0.549, including 49 and 28 physical
faults (\textbf{Fig. S10d-f}). On the N1-Ref timed windows the mean time
per decision fell from 39.4 to 19.1 s with glm-5.2 and from 36.2 to 18.6
s with deepseek-v4-pro; on N1-C, where the deferred windows included
most of the slow requests, it fell only from 43.3 to 42.3 s and from
32.1 to 30.2 s (\textbf{Fig. S10c}).

\textbf{Net1 by event class.} On the one-tank network Jev read the
bursts as normal transients (29 of 50) or sensor faults (20), recalling
0.02 of them against 0.92 for R1 and 0.84 for R2 transfer, and both
reviewers recalled 0.00-0.02; on N1-C no training-free method except R1
(0.85) recognized a growing leak (\textbf{Fig. S11c,d}). The pump trip
was recalled only by glm-5.2 (0.30). Jev in turn recalled every pump-off
attack spoofed as running (1.00) and 0.64 of the level-spoof-starve
attacks, which R2 transfer recalled at 0.52 and 0.12. With the evaluated
subtype removed from the C-Town labels, R2 fell to 0.300 on N1-Ref and
Jev beat it by +0.138 {[}0.084, 0.192{]} (\textbf{Fig. S13b}). The rule
cascade scored below R1 on both sets (\ensuremath{-}0.255 {[}\ensuremath{-}0.312, \ensuremath{-}0.200{]} and
\ensuremath{-}0.066 {[}\ensuremath{-}0.124, \ensuremath{-}0.011{]}). R2 refitted on Net1 labels overtook Jev
with one label per class on N1-Ref and with four on N1-C (\textbf{Fig.
S12}), so the events are separable in the evidence, but on this network
only by a model shown its own labels. A utility should therefore
confirm, on simulated events on its own network, that every event class
trips a check before it lets the gate settle windows; the redraw counts
of \textbf{Fig. S10b} are one such check.

\section*{Note S12. Secondary
comparisons}\label{note-s12.-secondary-comparisons}

The secondary, uncorrected comparisons of rounds 4-6 are collected in
\textbf{Fig. S13}, as differences in macro-F1 with paired,
class-stratified bootstrap 95\% intervals. Among the cascade comparisons
(\textbf{Fig. S13a}), the deepseek-v4-pro cascade differed from
deepseek-v4-pro alone by +0.010 {[}\ensuremath{-}0.016, 0.035{]} on E1c and \ensuremath{-}0.009
{[}\ensuremath{-}0.038, 0.018{]} on E4c, and the rule cascade scored below R1 on
every transfer set. Among the baseline comparisons (\textbf{Fig. S13b}),
Jev beat R2 LOSO on every set where it was defined (+0.417 on E1c,
+0.446 on N3-Ref, +0.138 on N1-Ref), and the rule cascade beat R2 LOSO
by +0.352 on E1c. Against R2 full, the difference was \ensuremath{-}0.243 and \ensuremath{-}0.237
for Jev on E1c and E4c, and \ensuremath{-}0.142 and \ensuremath{-}0.161 for the LLM cascade. The
comparisons with R2 own network are reported as a reference, not as
claims, because that model needs labelled data from the evaluated
network. The pre-registered primary tests of all rounds are listed in
\textbf{Table S3}.

\section*{Supplementary references}\label{supplementary-references}

Taormina, R., Galelli, S., Tippenhauer, N.O., Salomons, E., Ostfeld, A.,
Eliades, D.G., Aghashahi, M., Sundararajan, R., Pourahmadi, M., Banks,
M.K., Brentan, B.M., Campbell, E., Lima, G., Manzi, D., Ayala-Cabrera,
D., Herrera, M., Montalvo, I., Izquierdo, J., Luvizotto, E., Chandy,
S.E., Rasekh, A., Barker, Z.A., Campbell, B., Shafiee, M.E., Giacomoni,
M., Gatsis, N., Taha, A., Abokifa, A.A., Haddad, K., Lo, C.S., Biswas,
P., Pasha, M.F.K., Kc, B., Somasundaram, S.L., Housh, M., Ohar, Z.,
2018. Battle of the attack detection algorithms: disclosing cyber
attacks on water distribution networks. J. Water Resour. Plan. Manag.
144 (8), 04018048. https://doi.org/10.1061/(asce)wr.1943-5452.0000969

\section*{Tables}\label{tables}

\textbf{Table S1.} Splits, seed bases and use in the six rounds. Every
sealed set was evaluated once. The reference family is the
in-distribution subtype family; unseen families A, B and C (C, the
look-alike family) are the new families of rounds 1-2, 3 and 4-6.

\begin{longtable}[]{@{}
  >{\raggedright\arraybackslash}p{(\linewidth - 10\tabcolsep) * \real{0.0950}}
  >{\raggedright\arraybackslash}p{(\linewidth - 10\tabcolsep) * \real{0.1762}}
  >{\raggedright\arraybackslash}p{(\linewidth - 10\tabcolsep) * \real{0.1534}}
  >{\raggedright\arraybackslash}p{(\linewidth - 10\tabcolsep) * \real{0.1824}}
  >{\raggedright\arraybackslash}p{(\linewidth - 10\tabcolsep) * \real{0.1372}}
  >{\raggedright\arraybackslash}p{(\linewidth - 10\tabcolsep) * \real{0.2558}}@{}}
\toprule

\begin{minipage}[b]{\linewidth}\raggedright
Network
\end{minipage} & \begin{minipage}[b]{\linewidth}\raggedright
Split
\end{minipage} & \begin{minipage}[b]{\linewidth}\raggedright
Seed base
\end{minipage} & \begin{minipage}[b]{\linewidth}\raggedright
Windows
\end{minipage} & \begin{minipage}[b]{\linewidth}\raggedright
Family
\end{minipage} & \begin{minipage}[b]{\linewidth}\raggedright
Use
\end{minipage} \\
\midrule

\endhead
\bottomrule

\endlastfoot
C-Town & Training & 1,100,000 & 200 (50 per class) & Reference & R2
training; label-scarcity draws \\
C-Town & Development & 1,300,000 & 40 (10 per class) & Reference &
Question choice; unlabelled calibration prior \\
C-Town & Round 1 & 1,500,000/ 1,700,000 & 40 / 40 & Reference/ Unseen A
& Sealed E1 / E2 \\
C-Town & Round 2 & 5,050,000/ 5,060,000 & 200 / 200 & Reference/ Unseen
A & Sealed E1r2 / E2r2 \\
C-Town & Round 3 & 5,070,000/ 5,080,000 & 200 / 200 & Reference/ Unseen
B & Sealed E1r3 / E3 \\
C-Town & Round 4 & 5,090,000/ 5,095,000 & 200 / 200 & Reference/ Unseen
C & Sealed E1c / E4c \\
Net3 & Probe & 11,070,000 & - & Reference, Unseen C & Realisability of
subtypes; gates only \\
Net3 & Gate diagnostic & 12,000,000 & 30 & Reference bursts & Check
statistics only \\
Net3 & Round 5 training & 10,050,000 & 200 & Reference & R2 own network;
label-scarcity draws \\
Net3 & Round 5 & 10,010,000/ 10,020,000 & 200 / 200 & Reference/ Unseen
C & Sealed N3-Ref / N3-C \\
Net1 & Probe & 13,073,500 & - & Reference, Unseen C & Realisability of
subtypes; gates only \\
Net1 & Development & 200,000,000 & - & Reference, Unseen C & Design
checks; gates and R1 only \\
Net1 & Round 6 training & 13,063,500 & 200 & Reference & R2 own network;
label-scarcity draws \\
Net1 & Round 6 & 13,033,500/ 13,043,500 & 200 / 200 & Reference/ Unseen
C & Sealed N1-Ref / N1-C \\
\end{longtable}

\textbf{Table S2.} Rule tree R1. The rules are evaluated in order, and
the first rule that applies decides.

\begin{longtable}[]{@{}
  >{\raggedright\arraybackslash}p{(\linewidth - 4\tabcolsep) * \real{0.0620}}
  >{\raggedright\arraybackslash}p{(\linewidth - 4\tabcolsep) * \real{0.7586}}
  >{\raggedright\arraybackslash}p{(\linewidth - 4\tabcolsep) * \real{0.1794}}@{}}
\toprule

\begin{minipage}[b]{\linewidth}\raggedright
Rule
\end{minipage} & \begin{minipage}[b]{\linewidth}\raggedright
Condition
\end{minipage} & \begin{minipage}[b]{\linewidth}\raggedright
Decision
\end{minipage} \\
\midrule

\endhead
\bottomrule

\endlastfoot
1 & System demand violated and pressure pattern widespread & Normal
transient \\
2 & Pump head gain and tank level both violated & Cyberattack \\
3 & Tank mass balance violated and control logic violated &
Cyberattack \\
3\ensuremath{\prime} & Tank mass balance violated and control logic not violated & Sensor
fault \\
4 & Pump status versus flow violated, or exactly one anomalous channel
without a system-demand violation & Sensor fault \\
5 & System demand violated, or pressure pattern localized & Physical
fault \\
6 & Otherwise & Normal transient \\
\end{longtable}

\textbf{Table S3.} Every pre-registered primary test of rounds 1, 2,
2-LLM, 3, 4, 5 and 6. \ensuremath{\Delta} is the macro-F1 of the first method minus that
of the second on the same sealed windows, with a class-stratified paired
bootstrap of 10,000 resamples (95\% CI). Superiority p is one-sided and
non-inferiority p (margin 0.03) is the share of resamples at or below
\ensuremath{-}0.03 (\textbf{Eq. 6}); round 2-LLM uses two-sided p.~Holm's procedure
(\textbf{Eq. 7}) controls the family-wise error at \ensuremath{\alpha} = 0.05 within each
round (round 1: 4 tests; 2: 3; 2-LLM: 14; 3: 5; 4: 7; 5: 8, the
pre-registered family, \textbf{Note S11}; 6: 4); the Holm column gives
the step-down threshold, and Met means the null hypothesis was rejected.
``A \textgreater{} B'' is superiority of A over B, ``Jev \ensuremath{
e} M'' the
two-sided comparison with LLM M, and ``NI'' non-inferiority of the gated
LLM cascade to its reviewer alone.

\begin{longtable}[]{@{}
  >{\raggedright\arraybackslash}p{(\linewidth - 12\tabcolsep) * \real{0.0869}}
  >{\raggedright\arraybackslash}p{(\linewidth - 12\tabcolsep) * \real{0.1269}}
  >{\raggedright\arraybackslash}p{(\linewidth - 12\tabcolsep) * \real{0.2098}}
  >{\raggedright\arraybackslash}p{(\linewidth - 12\tabcolsep) * \real{0.2138}}
  >{\raggedright\arraybackslash}p{(\linewidth - 12\tabcolsep) * \real{0.1526}}
  >{\raggedright\arraybackslash}p{(\linewidth - 12\tabcolsep) * \real{0.1222}}
  >{\raggedright\arraybackslash}p{(\linewidth - 12\tabcolsep) * \real{0.0877}}@{}}
\toprule

\begin{minipage}[b]{\linewidth}\raggedright
Round
\end{minipage} & \begin{minipage}[b]{\linewidth}\raggedright
Set
\end{minipage} & \begin{minipage}[b]{\linewidth}\raggedright
Test
\end{minipage} & \begin{minipage}[b]{\linewidth}\raggedright
\ensuremath{\Delta} macro-F1 {[}95\% CI{]}
\end{minipage} & \begin{minipage}[b]{\linewidth}\raggedright
p
\end{minipage} & \begin{minipage}[b]{\linewidth}\raggedright
Holm
\end{minipage} & \begin{minipage}[b]{\linewidth}\raggedright
Met
\end{minipage} \\
\midrule

\endhead
\bottomrule

\endlastfoot
1 & E2 & Jev \textgreater{} R2 full & \ensuremath{-}0.057 {[}\ensuremath{-}0.198, 0.078{]} &
0.8055 & 0.0500 & no \\
1 & E2 & Jev \textgreater{} R1 & +0.144 {[}\ensuremath{-}0.044, 0.325{]} & 0.0676 &
0.0167 & no \\
1 & E1-LOSO & Jev \textgreater{} R2 LOSO & +0.361 {[}0.189, 0.523{]} &
\textless0.0001 & 0.0125 & yes \\
1 & E1-LOSO & Jev \textgreater{} R1 & +0.063 {[}\ensuremath{-}0.144, 0.262{]} &
0.2737 & 0.0250 & no \\
2 & E1r2 & Jev \textgreater{} R1 & +0.022 {[}\ensuremath{-}0.063, 0.107{]} & 0.3065 &
0.0250 & no \\
2 & E1r2-LOSO & Jev \textgreater{} R2 LOSO & +0.387 {[}0.318, 0.455{]} &
\textless0.0001 & 0.0167 & yes \\
2 & E2r2 & Jev \textgreater{} R1 & \ensuremath{-}0.044 {[}\ensuremath{-}0.116, 0.028{]} & 0.8874 &
0.0500 & no \\
2-LLM & E1r2 & Jev \ensuremath{
e} deepseek-v4-pro & \ensuremath{-}0.094 {[}\ensuremath{-}0.157, \ensuremath{-}0.033{]} &
0.0026 & 0.0045 & yes \\
2-LLM & E1r2 & Jev \ensuremath{
e} minimax-m2.7 & +0.021 {[}\ensuremath{-}0.046, 0.091{]} & 0.5440
& 0.0125 & no \\
2-LLM & E1r2 & Jev \ensuremath{
e} glm-5.2 & \ensuremath{-}0.078 {[}\ensuremath{-}0.133, \ensuremath{-}0.022{]} & 0.0066 &
0.0050 & no \\
2-LLM & E1r2 & Jev \ensuremath{
e} gemma4 & \ensuremath{-}0.013 {[}\ensuremath{-}0.087, 0.061{]} & 0.7170 &
0.0250 & no \\
2-LLM & E1r2 & Jev \ensuremath{
e} gemma4:31b & \ensuremath{-}0.003 {[}\ensuremath{-}0.078, 0.071{]} & 0.9226 &
0.0500 & no \\
2-LLM & E1r2 & Jev \ensuremath{
e} gpt-oss:20b & +0.076 {[}0.006, 0.144{]} & 0.0330 &
0.0063 & no \\
2-LLM & E1r2 & Jev \ensuremath{
e} gpt-oss:120b & +0.013 {[}\ensuremath{-}0.058, 0.087{]} & 0.7088
& 0.0167 & no \\
2-LLM & E2r2 & Jev \ensuremath{
e} deepseek-v4-pro & \ensuremath{-}0.203 {[}\ensuremath{-}0.278, \ensuremath{-}0.131{]} &
\textless0.0001 & 0.0036 & yes \\
2-LLM & E2r2 & Jev \ensuremath{
e} minimax-m2.7 & \ensuremath{-}0.044 {[}\ensuremath{-}0.114, 0.025{]} & 0.2234
& 0.0071 & no \\
2-LLM & E2r2 & Jev \ensuremath{
e} glm-5.2 & \ensuremath{-}0.264 {[}\ensuremath{-}0.337, \ensuremath{-}0.191{]} &
\textless0.0001 & 0.0038 & yes \\
2-LLM & E2r2 & Jev \ensuremath{
e} gemma4 & \ensuremath{-}0.036 {[}\ensuremath{-}0.102, 0.031{]} & 0.2898 &
0.0100 & no \\
2-LLM & E2r2 & Jev \ensuremath{
e} gemma4:31b & \ensuremath{-}0.041 {[}\ensuremath{-}0.110, 0.029{]} & 0.2484 &
0.0083 & no \\
2-LLM & E2r2 & Jev \ensuremath{
e} gpt-oss:20b & \ensuremath{-}0.098 {[}\ensuremath{-}0.175, \ensuremath{-}0.021{]} & 0.0108
& 0.0056 & no \\
2-LLM & E2r2 & Jev \ensuremath{
e} gpt-oss:120b & \ensuremath{-}0.134 {[}\ensuremath{-}0.208, \ensuremath{-}0.060{]} & 0.0008
& 0.0042 & yes \\
3 & E1r3 & Jev \textgreater{} R1 & +0.026 {[}\ensuremath{-}0.065, 0.117{]} & 0.2768 &
0.0125 & no \\
3 & E1r3 & Jev \textgreater{} R1b & \ensuremath{-}0.080 {[}\ensuremath{-}0.156, \ensuremath{-}0.003{]} & 0.9784
& 0.0500 & no \\
3 & E1r3-LOSO & Jev \textgreater{} R2 LOSO & +0.390 {[}0.321, 0.458{]} &
\textless0.0001 & 0.0100 & yes \\
3 & E3 & Jev \textgreater{} R1 & \ensuremath{-}0.064 {[}\ensuremath{-}0.147, 0.020{]} & 0.9347 &
0.0250 & no \\
3 & E3 & Jev \textgreater{} R1b & \ensuremath{-}0.051 {[}\ensuremath{-}0.139, 0.033{]} & 0.8857 &
0.0167 & no \\
4 & E1c & Rule cascade \textgreater{} R1 & \ensuremath{-}0.011 {[}\ensuremath{-}0.059, 0.037{]} &
0.6623 & 0.0250 & no \\
4 & E1c & Rule cascade \textgreater{} Jev & \ensuremath{-}0.064 {[}\ensuremath{-}0.130, 0.005{]} &
0.9649 & 0.0500 & no \\
4 & E1c & LLM cascade (glm-5.2) NI glm-5.2 & +0.011 {[}\ensuremath{-}0.011, 0.033{]}
& \textless0.0001 & 0.0083 & yes \\
4 & E1c-LOSO & Jev \textgreater{} R2 LOSO & +0.417 {[}0.344, 0.488{]} &
\textless0.0001 & 0.0100 & yes \\
4 & E4c & Rule cascade \textgreater{} R1 & +0.009 {[}\ensuremath{-}0.045, 0.059{]} &
0.3633 & 0.0125 & no \\
4 & E4c & Rule cascade \textgreater{} Jev & \ensuremath{-}0.006 {[}\ensuremath{-}0.063, 0.051{]} &
0.5680 & 0.0167 & no \\
4 & E4c & LLM cascade (glm-5.2) NI glm-5.2 & +0.016 {[}0.003, 0.033{]} &
\textless0.0001 & 0.0071 & yes \\
5 & N3-Ref & LLM cascade (glm-5.2) NI glm-5.2 & \ensuremath{-}0.004 {[}\ensuremath{-}0.013,
0.001{]} & \textless0.0001 & 0.0063 & yes \\
5 & N3-Ref & Jev \textgreater{} R2 transfer & \ensuremath{-}0.130 {[}\ensuremath{-}0.207,
\ensuremath{-}0.051{]} & 0.9993 & 0.0250 & no \\
5 & N3-C & LLM cascade (glm-5.2) NI glm-5.2 & \ensuremath{-}0.007 {[}\ensuremath{-}0.026, 0.008{]}
& 0.0117 & 0.0083 & no \\
5 & N3-C & Jev \textgreater{} R2 transfer & \ensuremath{-}0.112 {[}\ensuremath{-}0.184, \ensuremath{-}0.043{]}
& 0.9997 & 0.0500 & no \\
6 & N1-Ref & LLM cascade (glm-5.2) NI glm-5.2 & \ensuremath{-}0.002 {[}\ensuremath{-}0.016,
0.012{]} & 0.0002 & 0.0167 & yes \\
6 & N1-Ref & Jev \textgreater{} R2 transfer & \ensuremath{-}0.165 {[}\ensuremath{-}0.249,
\ensuremath{-}0.084{]} & 0.9999 & 0.0500 & no \\
6 & N1-C & LLM cascade (glm-5.2) NI glm-5.2 & \ensuremath{-}0.005 {[}\ensuremath{-}0.024, 0.009{]}
& 0.0079 & 0.0250 & yes \\
6 & N1-C & Jev \textgreater{} R2 transfer & +0.150 {[}0.086, 0.213{]} &
\textless0.0001 & 0.0125 & yes \\
\end{longtable}

\clearpage
\section*{Figures}\label{figures}

\begin{figure}[H]
\centering
\includegraphics[width=\linewidth,height=0.78\textheight,keepaspectratio]{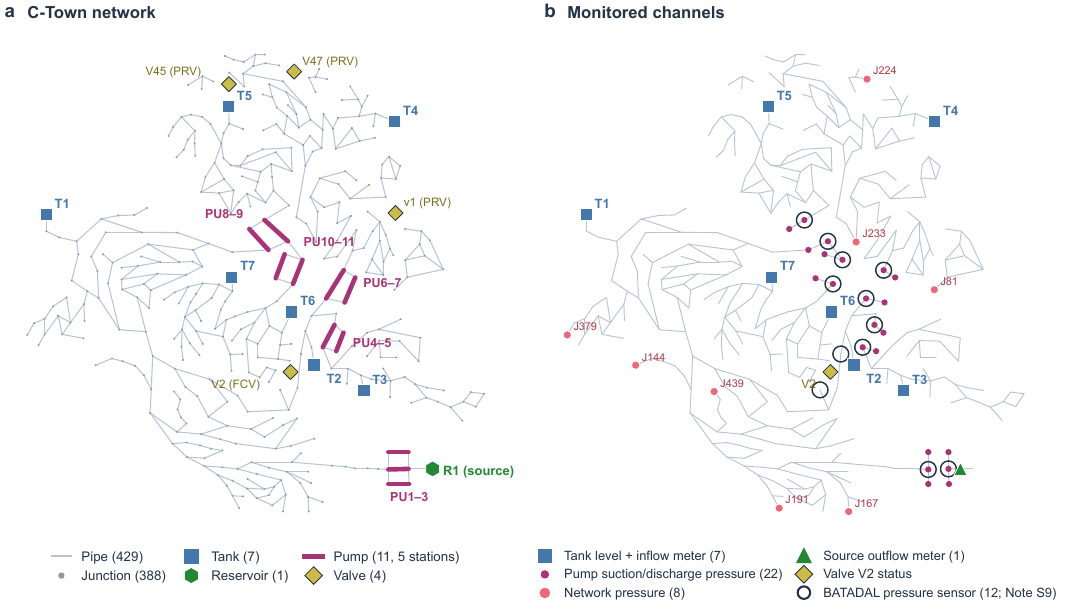}
\caption{C-Town network. (a) EPANET model used unchanged in rounds 1-4 and in the BATADAL study: 388 junctions, 429 pipes, 7 tanks, 1 reservoir, 11 pumps in five stations and 4 valves. (b) SCADA channels read by the evidence state: tank levels and inflow meters, pump flows and statuses, valve status, pump suction and discharge pressures, eight network pressures and the source outflow meter. Black rings mark the 12 pressure sensors of the BATADAL dataset.}
\label{fig:S1}
\end{figure}

\begin{figure}[H]
\centering
\includegraphics[width=\linewidth,height=0.78\textheight,keepaspectratio]{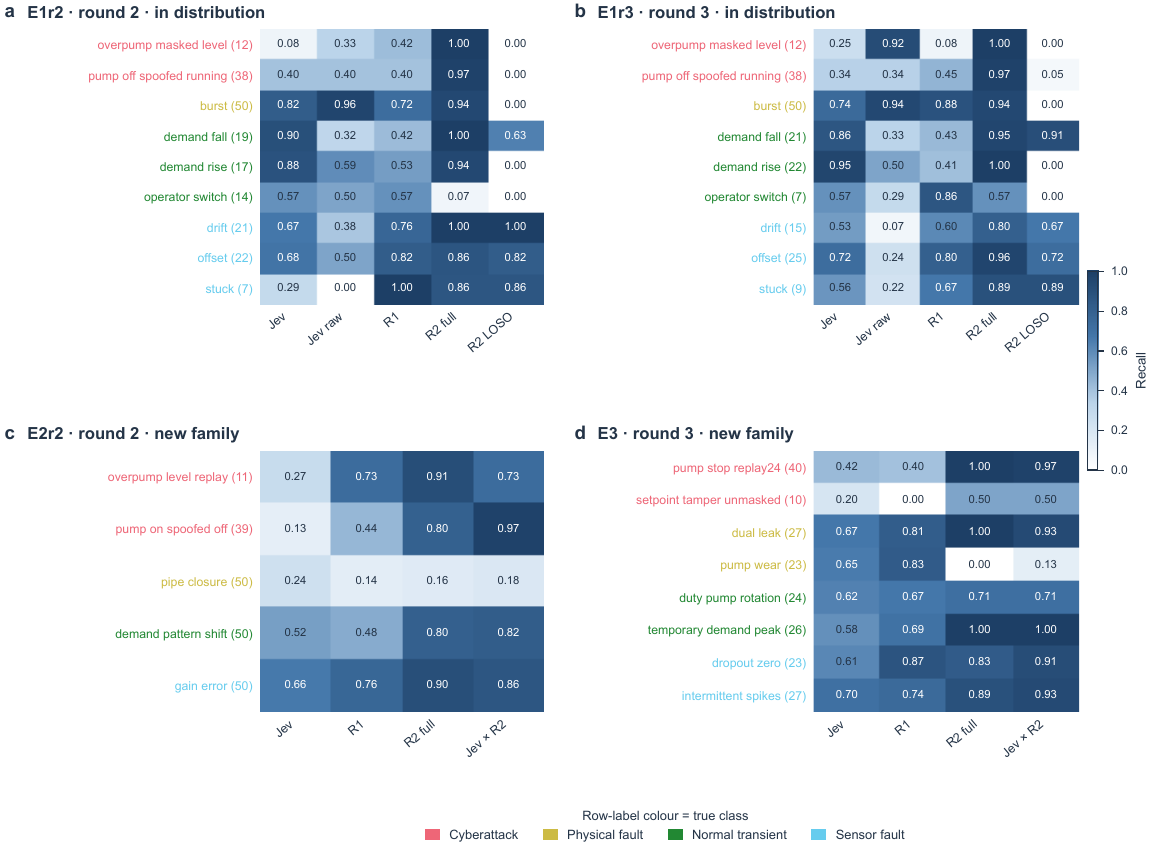}
\caption{Recall by subtype in rounds 2 and 3. (a) E1r2 and (b) E1r3 (in distribution) for calibrated Jev, raw Jev, R1, R2 full and R2 LOSO. (c) E2r2 and (d) E3 (new families, all subtypes unseen) for Jev, R1, R2 full and the product fusion of Jev and R2. Each cell is the recall of one subtype (window count in brackets); row-label colour gives the true class.}
\label{fig:S2}
\end{figure}

\begin{figure}[H]
\centering
\includegraphics[width=\linewidth,height=0.78\textheight,keepaspectratio]{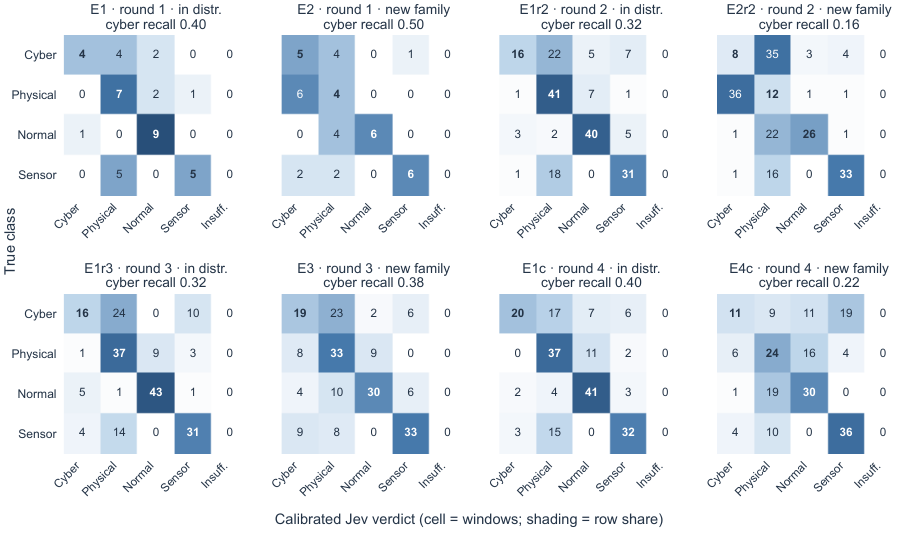}
\caption{Confusion matrices of calibrated Jev on the eight C-Town sealed sets. Rows are true classes and columns Jev verdicts; cells give window counts and shading the share of the row. Panel titles give the set, round, family and cyberattack recall.}
\label{fig:S3}
\end{figure}

\begin{figure}[H]
\centering
\includegraphics[width=\linewidth,height=0.78\textheight,keepaspectratio]{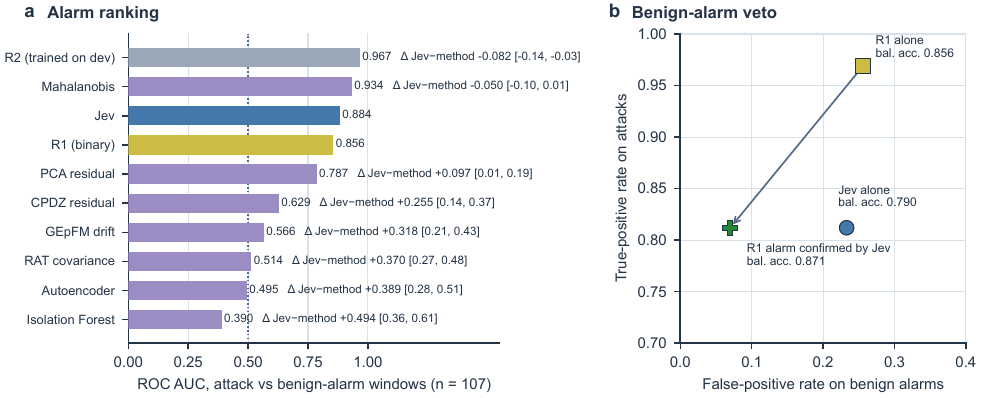}
\caption{Benign-alarm veto on BATADAL. (a) ROC AUC for separating attack windows from benign-alarm windows (n = 107), with the difference between Jev and each method and its bootstrap 95\% CI; the dotted line marks chance. (b) Operating points of R1 alone, Jev alone and the veto cascade in which Jev must confirm each R1 alarm, with their balanced accuracy.}
\label{fig:S4}
\end{figure}

\begin{figure}[H]
\centering
\includegraphics[width=\linewidth,height=0.78\textheight,keepaspectratio]{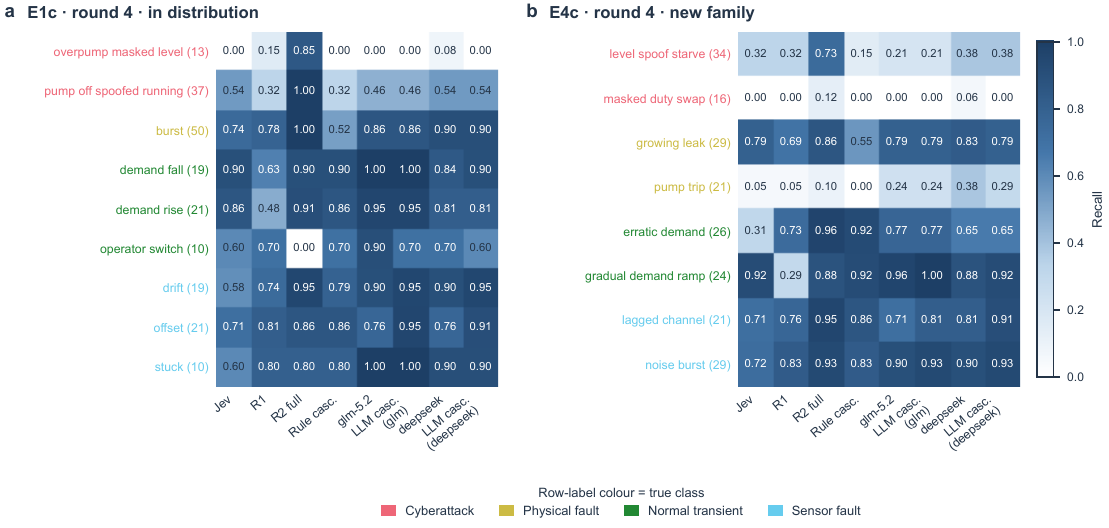}
\caption{Recall by subtype in round 4. (a) E1c (in distribution) and (b) E4c (look-alike family C) for Jev, R1, R2 full, the rule cascade, both reviewers alone and both LLM cascades. Row-label colour gives the true class and brackets the window count.}
\label{fig:S5}
\end{figure}

\begin{figure}[H]
\centering
\includegraphics[width=\linewidth,height=0.78\textheight,keepaspectratio]{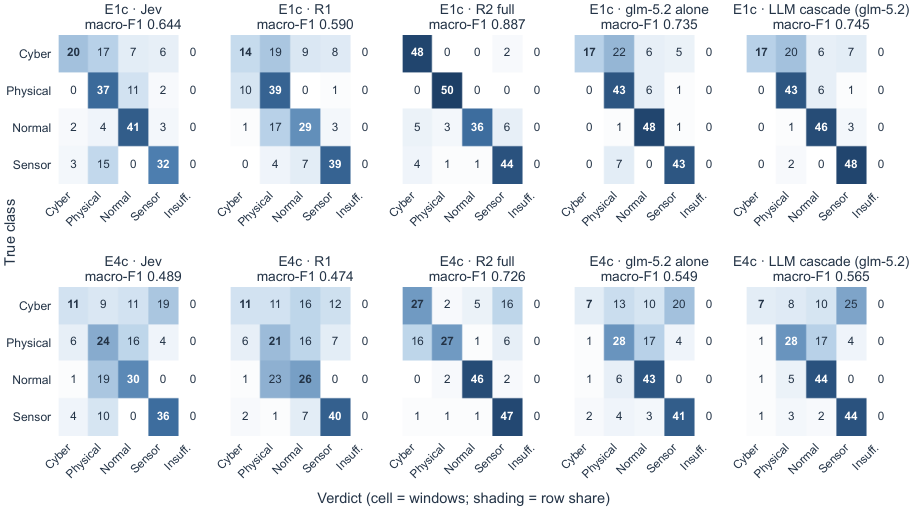}
\caption{Confusion matrices in round 4. E1c (top) and E4c (bottom) for Jev, R1, R2 full, glm-5.2 alone and the LLM cascade with glm-5.2 as reviewer. Cells give window counts and shading the share of the row; panel titles give the macro-F1.}
\label{fig:S6}
\end{figure}

\begin{figure}[H]
\centering
\includegraphics[width=\linewidth,height=0.78\textheight,keepaspectratio]{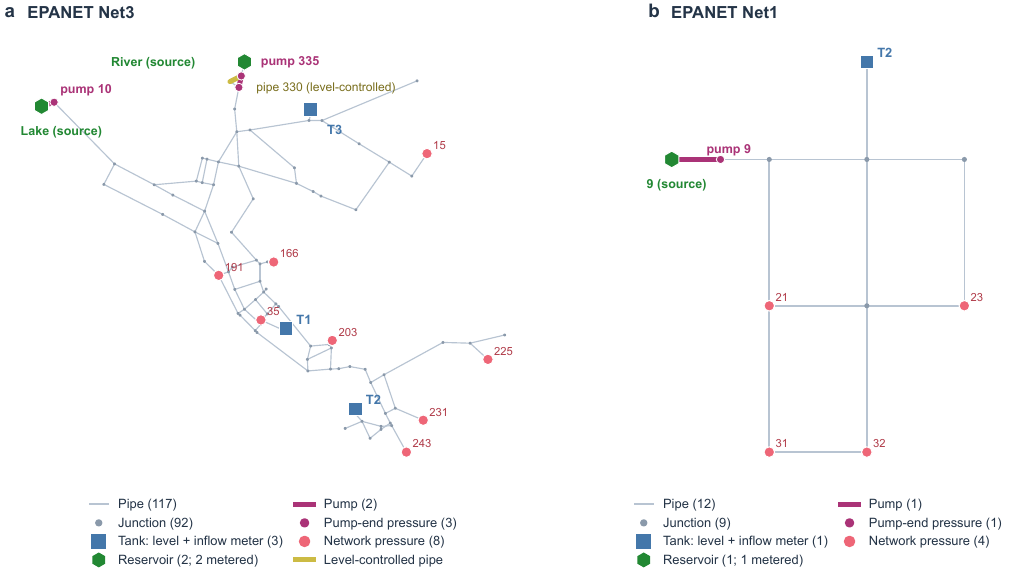}
\caption{Transfer networks. (a) EPANET Net3: 92 junctions, 117 pipes, 3 tanks, 2 metered reservoirs, 2 pumps and a pressure-zone bypass opened and closed by a tank level; 8 network pressures and 3 pump-end pressures are monitored. (b) EPANET Net1, with its demand patterns resampled to a 1-h step, used in round 6: 9 junctions, 12 pipes, 1 tank, 1 metered reservoir and 1 pump started and stopped by the tank level; 4 network pressures and 1 pump-end pressure are monitored. Every tank carries a level and an inflow meter.}
\label{fig:S7}
\end{figure}

\begin{figure}[H]
\centering
\includegraphics[width=\linewidth,height=0.78\textheight,keepaspectratio]{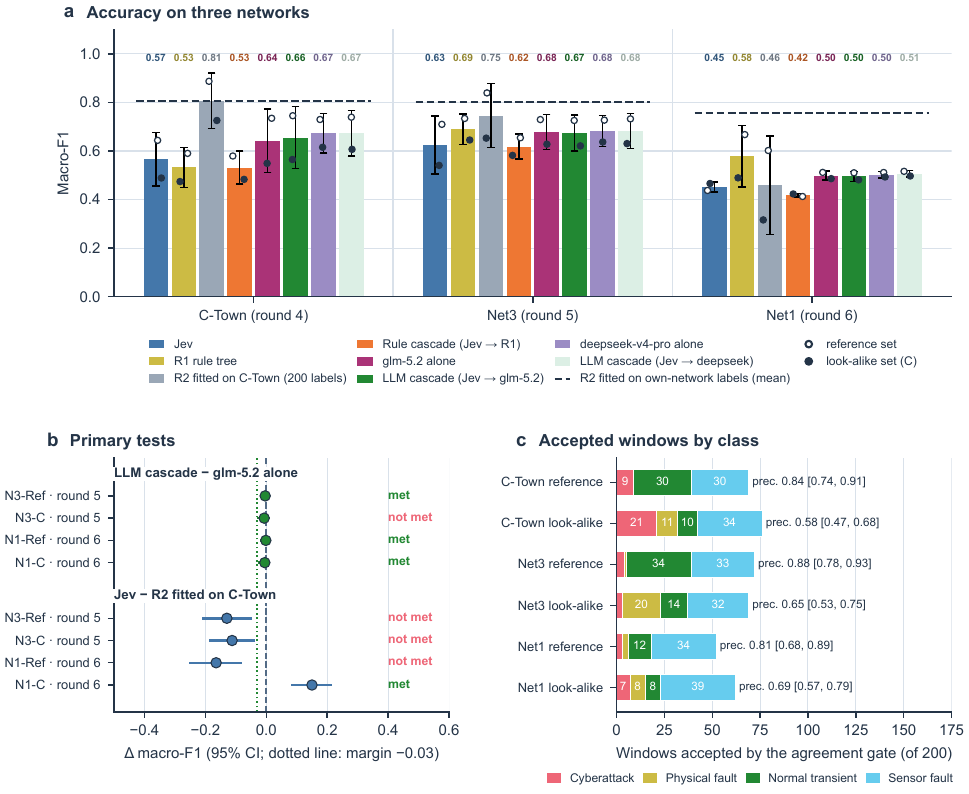}
\caption{Accuracy and primary tests across networks. (a) Macro-F1 on C-Town (round 4), Net3 (round 5) and Net1 (round 6). Bars show the mean of the reference-family and family-C sets (n = 2 per network), error bars \ensuremath{\pm}1 s.d. and dots the individual sets (hollow, reference family; filled, family C); the mean is printed above each bar. The dashed line is the mean of R2 fitted on the evaluated network's own labels. With n = 2 the s.d. only indicates the spread between the two sets. (b) The eight transfer primary tests: LLM cascade minus glm-5.2 alone (non-inferiority, margin \ensuremath{-}0.03, dotted) and Jev minus R2 transfer (superiority), with 95\% CIs. (c) Windows accepted by the agreement gate by true class, with accept precision and its Wilson 95\% interval (\textbf{Eq. S2}).}
\label{fig:S8}
\end{figure}

\begin{figure}[H]
\centering
\includegraphics[width=\linewidth,height=0.78\textheight,keepaspectratio]{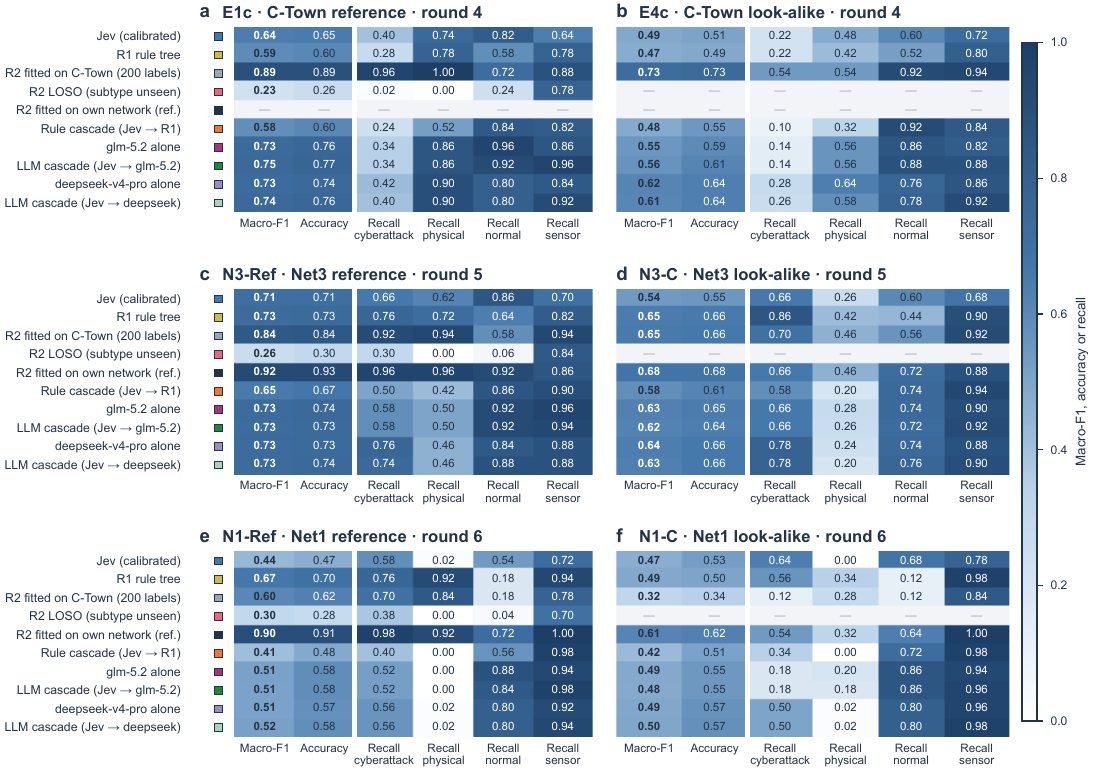}
\caption{All metrics on the fresh sealed sets of rounds 4-6. (a) E1c and (b) E4c (C-Town, round 4); (c) N3-Ref and (d) N3-C (Net3, round 5); (e) N1-Ref and (f) N1-C (Net1, round 6). Each row is a method and the columns give macro-F1, accuracy and per-class recall; on C-Town, R2 C-Town is R2 full, and a dash marks a method not defined for the set. R2 own network is a reference that needs labels of the evaluated network.}
\label{fig:S9}
\end{figure}

\begin{figure}[H]
\centering
\includegraphics[width=\linewidth,height=0.78\textheight,keepaspectratio]{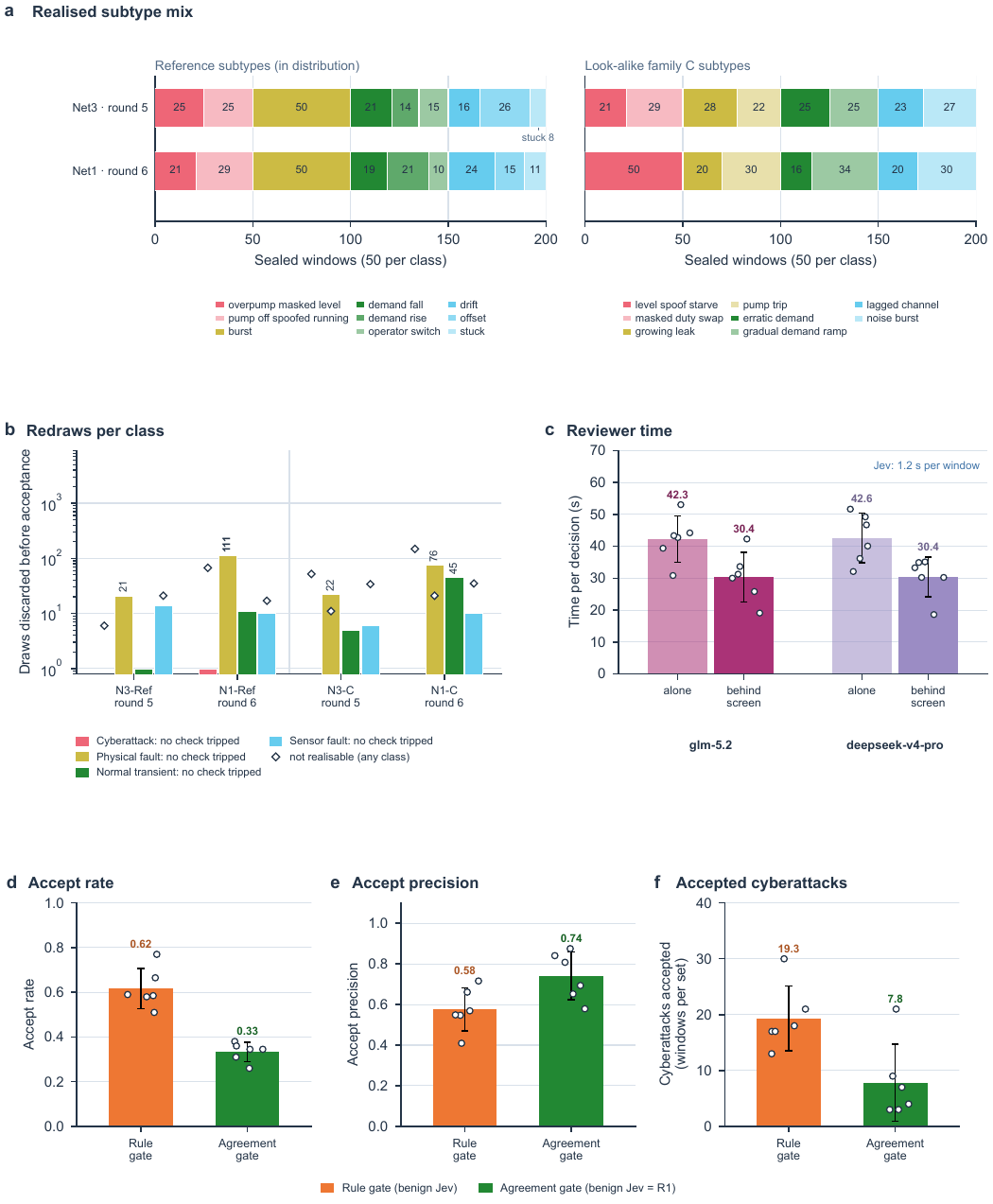}
\caption{Realised design, reviewer time and gate statistics. (a) Realised subtype mix of the transfer sets (50 windows per class). (b) Redraws per class before each accepted window (log scale): events that tripped no check (bars) and subtypes the network could not host (diamonds). (c) Mean reviewer time per decision alone and behind the agreement-gated screen on the windows timed one request at a time. (d) Accept rate, (e) accept precision and (f) accepted cyberattacks per set for the rule gate and the agreement gate. In c-f, bars show the mean over the six fresh sets (C-Town round 4, Net3 round 5 and Net1 round 6; n = 6), error bars \ensuremath{\pm}1 s.d. and white dots the individual sets; the mean is printed above each bar.}
\label{fig:S10}
\end{figure}

\begin{figure}[H]
\centering
\includegraphics[width=\linewidth,height=0.78\textheight,keepaspectratio]{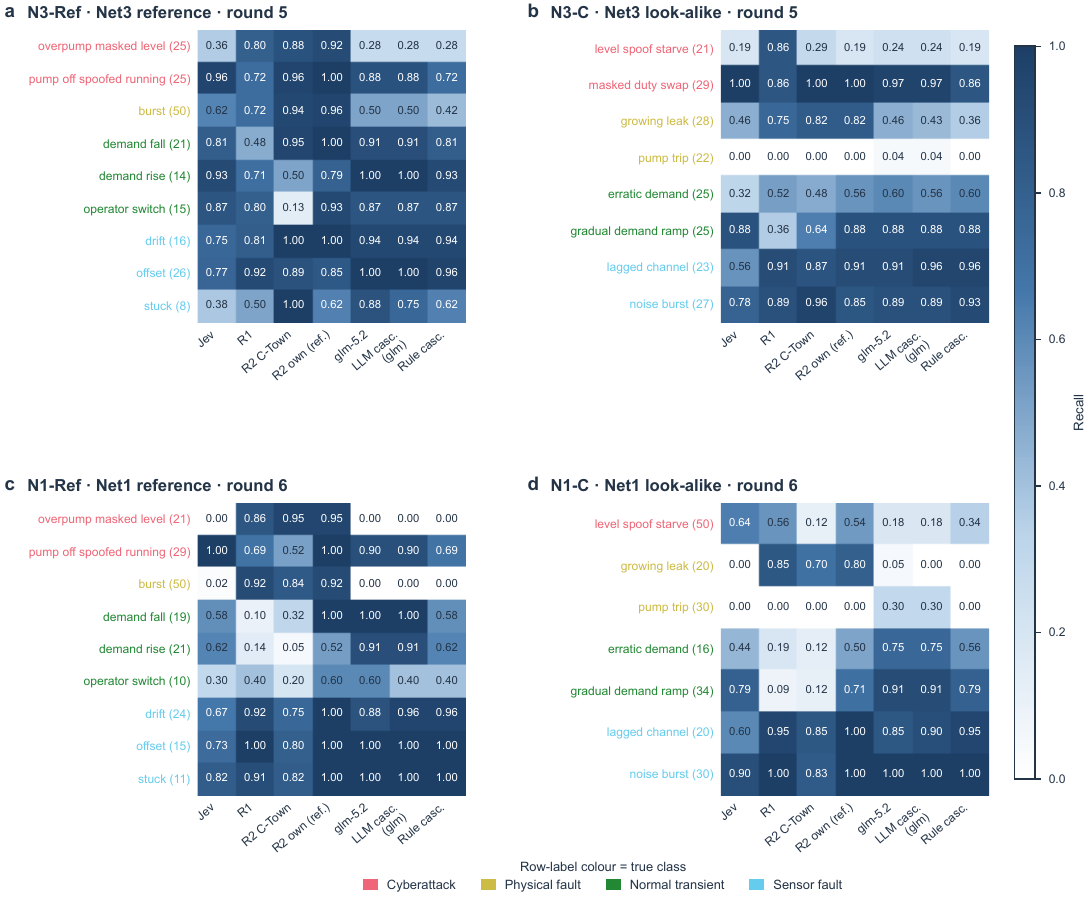}
\caption{Recall by subtype on the transfer sets. (a) N3-Ref and (b) N3-C (Net3, round 5); (c) N1-Ref and (d) N1-C (Net1, round 6). Columns are Jev, R1, R2 fitted on C-Town, R2 fitted on the evaluated network (reference), glm-5.2 alone, the LLM cascade with glm-5.2 and the rule cascade; row-label colour gives the true class and brackets the window count.}
\label{fig:S11}
\end{figure}

\begin{figure}[H]
\centering
\includegraphics[width=\linewidth,height=0.78\textheight,keepaspectratio]{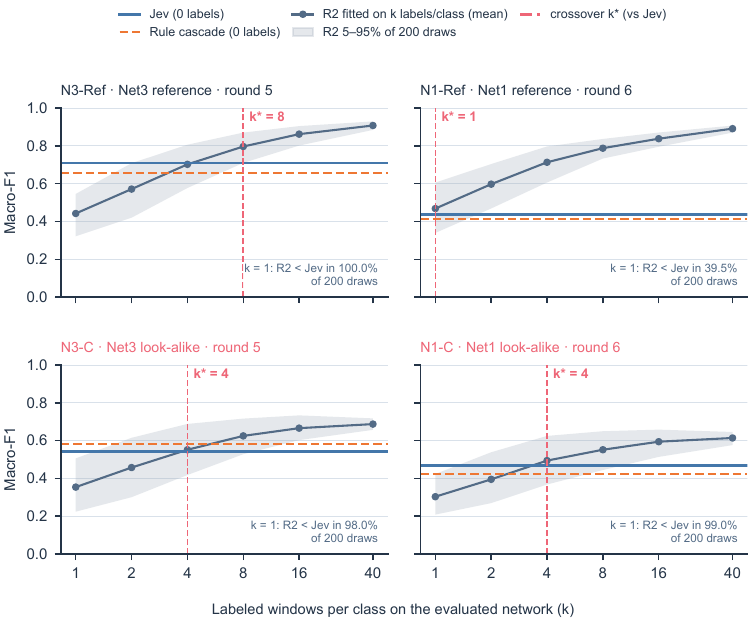}
\caption{Label scarcity on the transfer sets. Mean macro-F1 of R2 refitted on k labelled windows per class of the evaluated network (200 draws per k; shading, 5-95\% range of the draws), against Jev and the rule cascade, which use no labels. The dashed vertical line marks the crossover k* (\textbf{Eq. 8}), and the inset gives the share of draws at k = 1 in which R2 fell below Jev.}
\label{fig:S12}
\end{figure}

\begin{figure}[H]
\centering
\includegraphics[width=\linewidth,height=0.78\textheight,keepaspectratio]{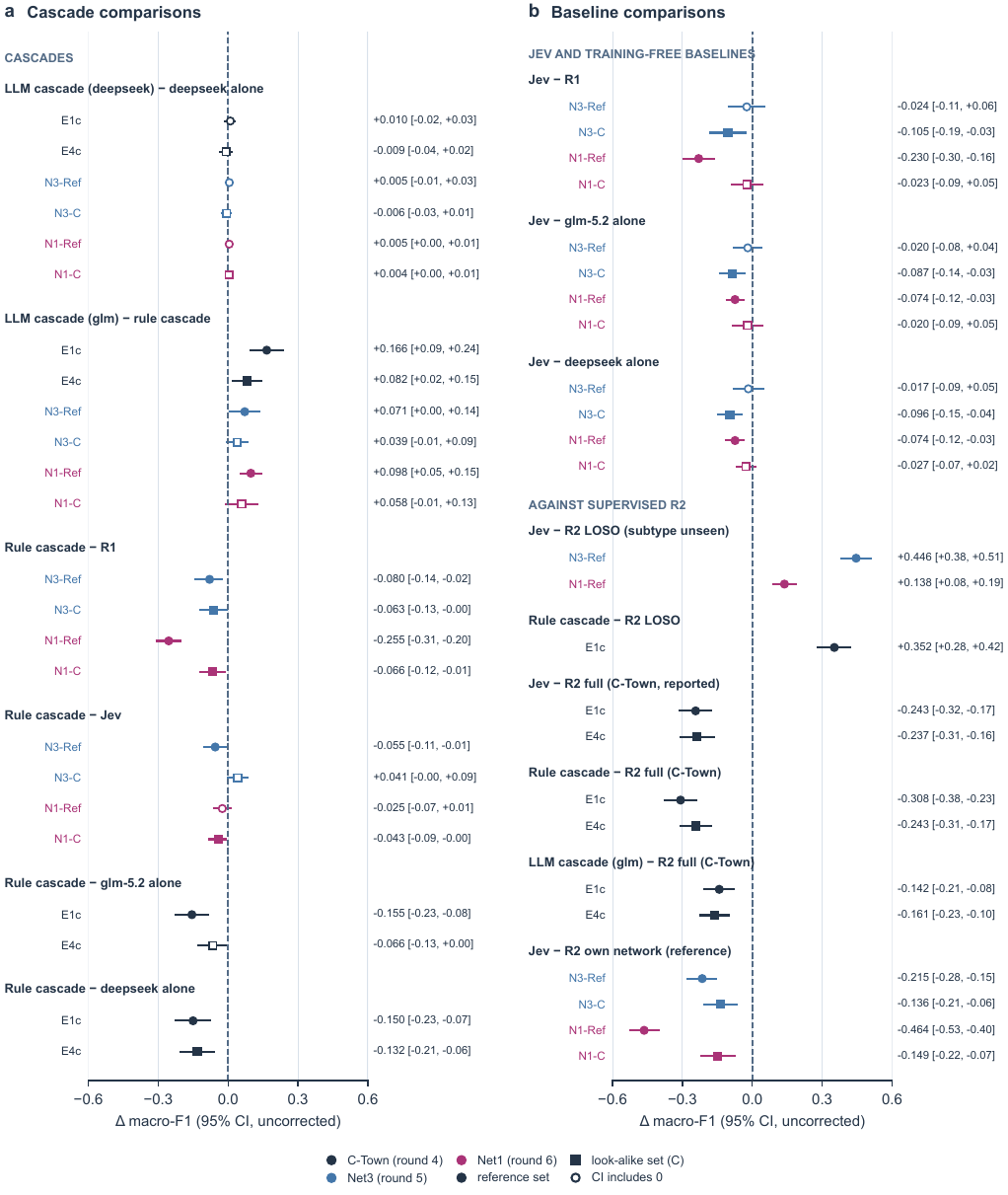}
\caption{Secondary comparisons of rounds 4-6. Differences in macro-F1 with paired, class-stratified bootstrap 95\% CIs, uncorrected for multiplicity; hollow markers indicate an interval that includes zero. (a) The cascades against their reviewers, their components and each other. (b) Jev and the cascades against R1, the LLMs alone, R2 LOSO, R2 full and R2 fitted on the evaluated network, the last reported as a reference. Colours give the network.}
\label{fig:S13}
\end{figure}

\end{document}